\pdfoutput=1 % Force pdfLaTeX on arXiv (all figures are PDF)
\documentclass[10pt,twocolumn,letterpaper]{article}

\usepackage[T1]{fontenc}       % T1 font encoding for robust PDF text output
\usepackage[pagenumbers]{cvpr} % To force page numbers, e.g. for an arXiv version

\newcommand{\TODO}[1]{\textbf{\color{red}[TODO: #1]}}
\newcommand{\ling}[1]{\textbf{\color{blue}[Ling: #1]}}
\newcommand{\bingchen}[1]{\textbf{\color{orange}[Bingchen: #1]}}
\newcommand{\amal}[1]{\textbf{\color{teal}[Amal: #1]}}
\newcommand{\maks}[1]{\textbf{\color{violet}[Maks: #1]}}
\renewcommand{\TODO}[1]{}

\renewcommand{\ling}[1]{}
\renewcommand{\bingchen}[1]{}
\renewcommand{\amal}[1]{}
\renewcommand{\maks}[1]{}

\definecolor{cvprblue}{rgb}{0.21,0.49,0.74}
\usepackage[pagebackref,breaklinks,colorlinks,allcolors=cvprblue]{hyperref}

\def\paperID{113} % *** Enter the Paper ID here
\def\confName{3DV\xspace}
\def\confYear{2027\xspace}

\title{3DHarnessBench: Probing Agentic 3D-to-Code Capabilities \\ of Frontier Vision-Language Models}

\author{
Ling Liu\textsuperscript{1,2,3}\quad
Bingchen Gong\textsuperscript{1,2} \quad
Amal Dev Parakkat\textsuperscript{1,3} \quad
Maks Ovsjanikov\textsuperscript{1,2}\\
\textsuperscript{1}Institut Polytechnique de Paris \\
\textsuperscript{2}École Polytechnique \quad
\textsuperscript{3}Télécom Paris\\
\href{https://llada60.github.io/3DHarnessBench/}{https://llada60.github.io/3DHarnessBench}
}

\begin{document}

\twocolumn[{%
    % Prevent \maketitle from starting another twocolumn block
    \renewcommand\twocolumn[1][]{##1}%
    \maketitle
    \centering
    \vspace{-1em}
    \includegraphics[width=\linewidth,trim={0 0 0 0},clip]{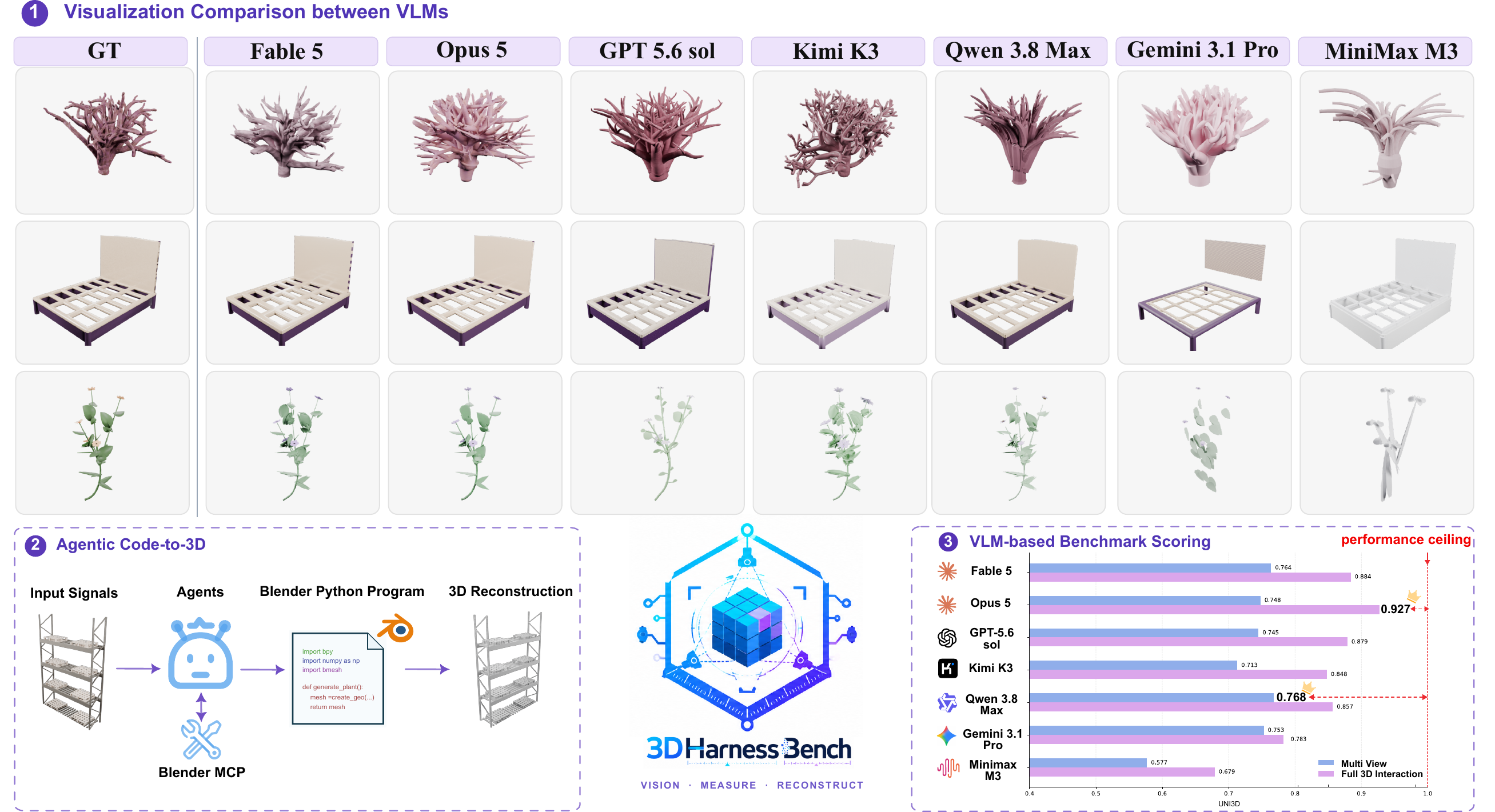}
    \captionof{figure}{\textbf{Overview of 3DHarnessBench.} We evaluate seven frontier VLM agents on reconstructing a target 3D object as executable Blender Python code. Panel 1: reconstructions from all seven agents under Full 3D Interaction, compared with the ground truth (GT). Panel 2: the agentic reconstruction loop, in which an agent converts target observations into a Blender Python program through Blender MCP and iteratively refines it. Panel 3: Uni3D feature similarity under Multi-view and Full 3D Interaction; richer target access improves every agent, yet the best score (0.927) still leaves a clear gap to the performance ceiling, and large cross-agent differences remain. 
    \vspace{1em}}
    \label{fig:teaser}
}]

\begin{abstract}
We introduce 3DHarnessBench, a benchmark that evaluates the agentic ability of frontier vision-language models (VLMs) to recover 3D geometry as Blender Python code from a variety of inputs. Unlike previous frameworks that prompt the VLMs with a \emph{fixed input} (e.g., a single rendering or a text description), 3DHarnessBench evaluates four separate harness settings that progressively enable \emph{active agentic exploration}, facilitated by recent Blender MCP functionality. Our hierarchy 
from Single-view, Multi-view, Active Visual (arbitrary viewpoint access), and Full 3D Interaction (complete access to the target object through Blender function calls) probes the models' abilities in both visual \emph{perception} and \emph{active inference, tool calling, and self-correction}.
We observe that the ability of all frontier models to recover 3D geometry improves significantly with richer function call access, although the improvements are strongly model-dependent revealing highly uneven agentic 3D-to-code capabilities. We will release the benchmark, code, outputs, and agent trajectories for reproducible 3D evaluation.

\vspace{0.1em} \noindent\makebox[\linewidth][c]{
\textit{`What I cannot create, I do not understand.''}
}\par \vspace{0.1em} \noindent\hfill \textit{\rule{1.8em}{0.4pt}\hspace{0.35em}Richard P. Feynman}

\end{abstract}

\section{Introduction}
\label{sec:intro}
Frontier vision-language models (VLMs) increasingly operate as agents: they call tools, render intermediate results, and can revise their own artifacts throughout long interactions. 
This raises a basic question about how much \textit{3D awareness} state-of-the-art VLMs can have when they act as agents. Specifically, are they able to incorporate additional information through tool calls, perform self-correction, and integrate both visual and geometric measurements into a coherent representation?
Prior evaluations of 3D competence largely probe for \emph{perceptual} understanding through recognition, question answering, or spatial-reasoning tasks~\cite{hong20233d,xu2024pointllm,zhu2024llava,zhang2026multiview}. Such probes evaluate whether a model can answer queries about 3D structure, but not whether it can actively \textit{generate} it. 

More recently, works in executable inverse graphics~\cite{sun20253d,kulits2024re,chen2025img2cad,he2026thinking}, propose to evaluate 3D capabilities \emph{constructively} by evaluating the model's ability to recover a Blender Python program whose execution produces the target 3D reconstruction. However, existing benchmarks for programmatic 3D
modeling typically provide \textit{a fixed input} to the VLM, e.g., a text prompt or a fixed
set of
renders~\cite{gao2026_3dcodebench,gu2025blendergym,ji2026codegen,du2024blenderllm,yang2026p3d},
even though modern VLM agents can control cameras, render scenes, and query
geometry through interactive 3D environments such as Blender
MCP~\cite{blender_mcp_server,ahujasid_blender_mcp,hu2024scenecraft,huang2024blenderalchemy,yin2026viga}.
We argue that a single prescribed protocol leaves the sources of failure entangled: a poor reconstruction may reflect weak image-3D inference, an inability to integrate multiple views~\cite{zhang2026multiview}, or simply the absence of evidence the model would have needed. Nor does it test whether an agent can \emph{acquire} useful evidence on its own, arguably the defining capability of an agent.

\begin{figure*}[t]
    \centering
    \includegraphics[width=\linewidth]{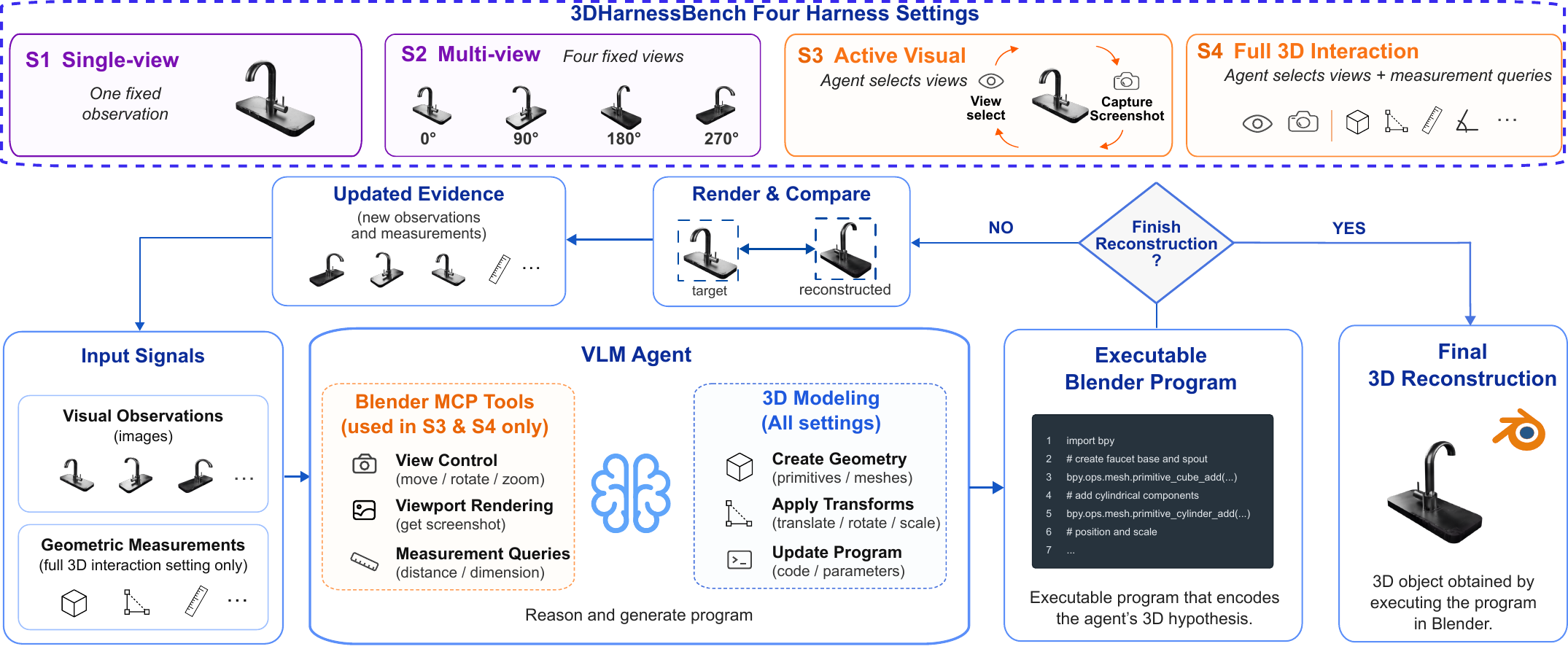}
    \caption{\textbf{Agentic 3D Reconstruction Pipeline.} The top panel shows four harness settings that progressively expand target access from a single fixed view (S1) to four fixed views (S2), agent-directed visual exploration (S3), and Full 3D Interaction with viewpoint control and geometric measurement queries (S4). The bottom panel shows the reconstruction pipeline: a VLM agent receives target-side evidence, updates its reconstruction as executable Blender Python code, renders the current result, and iteratively refines it as new observations or measurements become available. Blender MCP tools are available only in S3 and S4; the harness controls what evidence the agent can acquire, while the reconstruction objective remains fixed. \bingchen{added tool-availability note and takeaway}}
    \label{fig:pipeline}
\end{figure*}

To address these limitations, we introduce \textbf{3DHarnessBench}, a benchmark that keeps the executable reconstruction objective fixed while systematically varying the agent's access to the target (Fig.~\ref{fig:teaser}).
\bingchen{swapped figure citations so Fig. 1 is referenced before Fig. 2}
We consider four harnesses with increasing complexity (Fig.~\ref{fig:pipeline}): \emph{Single-view} provides one fixed image; \emph{Multi-view} provides four fixed views; \emph{Active Visual} lets the agent control the camera, inspecting the target from any viewpoint as its reconstruction evolves; and \emph{Full 3D Interaction} enables full Blender functionality, providing geometric queries such as bounding boxes, dimensions, and part statistics. 
These levels isolate VLMs' capabilities: 1) 3D inference from a single image, 2) integration of \textit{fixed} multi-view evidence, 3) active viewpoint selection, 4) metric grounding, and interactive self-refinement. To our knowledge, 3DHarnessBench is the first benchmark to probe 3D-to-code reconstruction under controlled levels of target-side 3D access. 
We instantiate it with 100
diverse objects derived from 3DCodeBench~\cite{gao2026_3dcodebench},
reusing its targets while replacing its fixed input protocol with our four
levels of target access, \bingchen{added the explicit 3DCodeBench contrast here (your point 5)} and
evaluate seven frontier VLMs, spanning proprietary and open-weight models, in
their official agent CLIs, connected to Blender MCP, so that each
model uses its native tool-calling and context-management machinery.
Reconstructions are scored on appearance, geometry, and topology, alongside
the interaction cost incurred under each harness.

The evaluation reveals a consistent access hierarchy with strongly
model-dependent improvements (Fig.~\ref{fig:teaser}).
First, passive multi-view evidence helps: moving from one to four fixed views yields lower Chamfer distance and higher Uni3D similarity for all seven agents. Second, Active Visual is the sharpest discriminator: several agents convert self-directed inspection into large gains, while others fall below their own fixed-view baselines. Moving from Multi-view to Active Visual shifts Uni3D by up to $+16\%$ for the strongest agents yet by as much as $-17\%$ for the weakest, widening the best-to-worst agent gap to $45\%$ of the top score, versus $25$--$33\%$ under fixed views; controlled ablations trace much of this gap to how agents handle their visual observation interface rather than to 3D reasoning itself.
Third, Full 3D Interaction improves every agent over its Single-view
baseline and yields the strongest geometry across the board, yet a substantial cross-model spread remains, indicating a wide range of truly agentic capabilities. 
%Finally, separating each agent's first executable reconstruction from its final one shows that forming a good initial 3D hypothesis and improving it through interaction are distinct capabilities.

Our contributions are threefold:
\begin{itemize}
	\item We formulate agentic 3D-to-code evaluation under controlled target access that progressively expands how an agent may observe,
	explore, and measure the target.
	\item We evaluate frontier
	VLMs in their official agent CLIs against Blender MCP environments, with four target-side
	access scenarios, and score them on metrics spanning appearance, geometry, topology, and
	interaction cost.
	\item We analyze seven frontier VLMs through controlled ablations on camera-pose cues, observation normalization, and measurement granularity, revealing an emerging but uneven ability to acquire and exploit 3D evidence actively. We will release the benchmark, code, outputs, and agent trajectories.
\end{itemize}

\section{Related Work}
\nbf{Probing 3D Competence in Vision-Language Models}
Recent work studies 3D understanding in multimodal language models from two
directions. One improves 3D reasoning through explicit 3D representations or
geometric supervision. 3D-LLM~\cite{hong20233d},
PointLLM~\cite{xu2024pointllm}, and LLaVA-3D~\cite{zhu2024llava}
incorporate 3D scenes or point clouds, while
SpatialVLM~\cite{chen2024spatialvlm} and
SpatialLLM~\cite{ma2025spatialllm} introduce spatially grounded supervision.
Other methods use native 3D grounding~\cite{wang2025n3d}, predictive spatial
representations~\cite{jiang2026spa3r}, or reconstruction-grounded
reasoning~\cite{hu2026g,fan2026vlm}. A complementary direction probes the
3D competence already present in general-purpose VLMs through spatial
reasoning, limited-view inference, and multi-view integration
tasks~\cite{liu20233daxiesprompts,ma20253dsrbench,yang2025thinking,
daxberger2025mm,yin2025spatial,yang2026mmsi,li2025viewspatial,
zhang2026revsi,zhang2026multiview}. These evaluations mainly measure spatial
judgments, metric reasoning, or question answering. 3DHarnessBench instead
uses \emph{3D reconstruction as the probe}: a frontier VLM is tasked with producing Blender Python code to generate the 3D structure, making errors in geometry,
topology, and appearance directly measurable.

\nbf{3D Reconstruction and Executable Inverse Graphics}
One line of 3D reconstruction develops image-conditioned models that directly
predict 3D representations~\cite{hong2024lrm,tochilkin2024triposr,
xu2024instantmesh,xiang2025structured,li2025triposg,chen2026sam}.
These methods primarily optimize reconstruction quality with dedicated 3D
architectures. Another line formulates reconstruction as structured or
executable inverse graphics. 3D-GPT~\cite{sun20253d}, Re-Thinking Inverse
Graphics~\cite{kulits2024re}, Img2CAD~\cite{chen2025img2cad}, and
MeshCoder~\cite{dai2026meshcoder} generate structured or programmatic 3D
representations, while 3D-CoS~\cite{wang20263d}, Thinking in
Blender~\cite{he2026thinking}, and VIGA~\cite{yin2026viga} combine VLM
reasoning, code generation, and visual feedback for 3D reconstruction.
Executable programs expose decisions about primitives, dimensions,
transformations, and part relationships while supporting iterative
refinement. 3DHarnessBench adopts this representation primarily as an
\emph{evaluation interface}, allowing the agent's inferred 3D structure to
be externalized as an executable program and measured as progressively
richer evidence becomes available.

\nbf{Benchmarks for Programmatic 3D Modeling}
Most closely related to our work are recent benchmarks that evaluate VLMs through executable 3D modeling and reconstruction. 3DCodeBench~\cite{gao2026_3dcodebench} studies
image-conditioned modeling in Blender, while
BlenderGym~\cite{gu2025blendergym} evaluates Blender program editing.
CodeGen-3D~\cite{ji2026codegen}, P3D-Bench~\cite{yang2026p3d},
VoxelCodeBench~\cite{zheng2026voxelcodebench}, and
Nova3D~\cite{noor2026nova3d} further study code-based or iterative 3D
generation. Related CAD benchmarks emphasize structural, parametric, and
functional reconstruction~\cite{du2024blenderllm,mallis2026text,
zhang2026benchcad,doris2026cadbench,wang2026text2cad,dong2026muse,
chen2026unicad}. Among these, 3DCodeBench is most closely related to our benchmark; however, it evaluates reconstruction only under \textit{fixed-image input settings}.
3DHarnessBench extends this by considering agentic Blender MCP interaction, enabling
agent-directed visual exploration and explicit geometric queries.

\nbf{Agentic 3D Reconstruction}
Active reconstruction studies how agents acquire informative observations
during 3D reconstruction. Prior work~\cite{feng2024naruto,
jiang2024fisherrf,yin2026viga,xu2026area3d} selects views using VLM guidance
or reconstruction uncertainty. In parallel, agentic 3D systems use iterative
tool interaction for content creation and
editing~\cite{hu2024scenecraft,huang2024blenderalchemy,lu2025ll3m,
wang2026ezblender,hayashi20253dify,guo2026worldclaw}. We also note
\textit{concurrent work} SceneActBench~\cite{zhao2026sceneactbench} which also evaluates interactive 3D agents through Blender MCP~\cite{blender_mcp_server,ahujasid_blender_mcp}. Unlike all of these, however, 3DHarnessBench focuses on \emph{3D reconstruction competence} under progressively richer target-side access, probing whether frontier VLM agents can acquire, interpret, and exploit visual and geometric evidence to construct and refine an executable
reconstruction.
\section{3DHarnessBench}
\label{sec:3dharnessbench}

3DHarnessBench probes the agentic 3D-to-code capabilities of frontier
vision-language models through executable object reconstruction. Given access
to an underlying 3D target through a prescribed interaction harness, an agent
must perceive the target, infer its 3D structure, and reconstruct it in Blender
by generating executable Python code. We progressively expand target access
from passive visual observations to agent-directed visual exploration and
explicit geometric interaction.

\subsection{Benchmark Formulation}
\label{sec:benchmark_formulation}

Let $G^\star$ denote a target 3D object. We evaluate a VLM agent under four harness configurations,
\begin{equation} \mathcal{S}_{\mathrm{single}} \prec \mathcal{S}_{\mathrm{multi}} \prec \mathcal{S}_{\mathrm{active}} \prec \mathcal{S}_{\mathrm{full}}, \end{equation}
where $\prec$ denotes increasingly informative access to evidence about $G^\star$. 

% Here, \emph{3D-to-code} means reconstructing $G^\star$ as an executable Blender Python program from fixed images, actively acquired visual observations, or explicit geometric information.

For each harness $s\in\mathcal{S}$, its interaction interface is
\begin{equation}
\mathcal{I}^{(s)}
=
\left(
\mathcal{U}^{(s)},
\mathcal{T}^{(s)},
\Omega^{(s)}
\right),
\end{equation}
where $\mathcal{U}^{(s)}$, $\mathcal{T}^{(s)}$, and $\Omega^{(s)}$ denote the task instruction, available tool space, and observation space, respectively. Thus, $\mathcal{I}^{(s)}$ determines what evidence about $G^\star$ the agent can acquire and how it can interact with the target and its current reconstruction.

At interaction step $t$, the agent conditions on the history $H_t^{(s)}=\left(
u_0,o_0,a_0,o_1,\ldots,a_{t-1},o_t \right)$, and selects the next tool invocation according to
\begin{equation}
a_t
\sim
\pi_\theta(\cdot\mid H_t^{(s)}).
\end{equation}
where $a_t \in \mathcal{T}^{(s)}$ \bingchen{renamed tool invocations from $t_t$ to $a_t$: $t$ was overloaded as the step index} and
$o_t \in \Omega^{(s)}$. Depending on the harness, the agent may execute and
render its current reconstruction, inspect the target from a new viewpoint,
or query geometric properties.

The agent iteratively updates a Blender Python program $P_t$. At the final interaction step $T$,
\begin{equation}
    P^{(s)} = P_T,
    \qquad
    \hat{G}^{(s)}
    =
    \mathcal{E}\!\left(P^{(s)}\right),
\end{equation}
where $\mathcal{E}$ denotes Blender execution and $\hat{G}^{(s)}$ is the final reconstruction. The executable program externalizes the agent's estimate of the target structure through explicit geometry, dimensions, transformations, spatial relationships, and appearance.
% ; evaluation is performed on the reconstructed object produced by executing the program rather than on a textual description.

As summarized in Figure~\ref{fig:pipeline}, the agent repeatedly acquires target evidence, updates its executable reconstruction, and refines it through tool interaction. The harness varies the accessible evidence while keeping the reconstruction objective and $G^\star$ fixed. By varying $\mathcal{I}^{(s)}$, 3DHarnessBench characterizes how frontier VLM agents exploit increasingly rich 3D evidence for executable reconstruction.

\subsection{Target Access Settings}
\label{sec:target_access}

The four harnesses form a progression from passive observation to active
visual exploration and, finally, explicit 3D interaction. They differ in the tools and observations available to the agent while sharing the
same executable reconstruction objective.

\subsubsection{Single-view and Multi-view}
\label{sec:fixed_view}

Single-view and Multi-view evaluate reconstruction from passive visual
evidence. In both settings, all target observations are prescribed in advance. They share the same
reconstruction-side tool space:
\begin{equation}
    \mathcal{T}^{(\mathrm{single})}
    =
    \mathcal{T}^{(\mathrm{multi})}
    =
    \mathcal{T}_{\mathrm{rec}},
\end{equation}
where $\mathcal{T}_{\mathrm{rec}}$ denotes tools for executing the generated
Blender program and rendering the current reconstruction.

At step $t$, the observations available to the agent are
\begin{equation}
    o_t^{(\mathrm{single})}
    =
    \{I^{\mathrm{tar}}, I_t^{\mathrm{rec}}\},
    \quad
    o_t^{(\mathrm{multi})}
    =
    \left\{
        \{I_v^{\mathrm{tar}}\}_{v=1}^{4},
        I_t^{\mathrm{rec}}
    \right\},
\end{equation}
where $I^{\mathrm{tar}}$ is the fixed target image in Single-view,
$I_v^{\mathrm{tar}}$ is the $v$-th fixed target view in Multi-view, and
$I_t^{\mathrm{rec}}$ is a rendered observation of the agent's current
reconstruction.

Single-view tests how much 3D structure an agent can infer from a single projection, in which substantial portions of the geometry may remain occluded or ambiguous. Multi-view reduces this ambiguity by providing four fixed views of the same object while preserving a passive observation setting. In both cases, the available visual evidence remains fixed throughout the reconstruction process.
% : the agent cannot acquire additional views, inspect regions according to its own uncertainty, or query geometric properties of the target.

\subsubsection{Active Visual}
\label{sec:active_visual}

Active Visual introduces agent-directed perception. Rather than receiving only
a fixed set of target views, the agent can decide how to inspect the underlying target 3D shape
during its reconstruction. Intuitively, this mimics a human modeler's ability to consider the reference subject from different views, while trying to recreate it. The tool space is
\begin{equation}
    \mathcal{T}^{(\mathrm{active})}
    =
    \mathcal{T}_{\mathrm{rec}}
    \cup
    \mathcal{T}_{\mathrm{view}},
\end{equation}
where $\mathcal{T}_{\mathrm{view}}$ provides target-side camera and viewport
control. The corresponding observation at step $t$ is
\begin{equation}
    o_t^{(\mathrm{active})}
    =
    \{I_t^{\mathrm{tar}}, I_t^{\mathrm{rec}}\},
\end{equation}
where $I_t^{\mathrm{tar}}$ is a target view acquired under an
agent-selected camera configuration.

The agent may change viewing direction, inspect previously unseen regions,
zoom into local details, zoom out to assess global proportions, and revisit
ambiguous structures as its reconstruction evolves. Target observation
thus becomes part of the agent's decision process. %: the agent must
%determine what visual evidence would be useful to acquire next.

After acquiring new observations, the agent may update and execute its Blender program, with further inspection and comparison as needed. Active Visual thus probes whether frontier VLM agents can strategically acquire and leverage visual evidence to iteratively refine their reconstructions.

\subsubsection{Full 3D Interaction}
\label{sec:full_access}

Full 3D Interaction extends agent-directed visual exploration with direct
access to the target object through unconstrained Blender MCP function
calls~\cite{blender_mcp_server}. These tools support Blender code execution, scene and object inspection, viewport control, screenshots, rendering, and geometric queries. Its tool space is
\begin{equation}
    \mathcal{T}^{(\mathrm{full})}
    =
    \mathcal{T}^{(\mathrm{active})}
    \cup
    \mathcal{T}_{\mathrm{measure}},
\end{equation}
where $\mathcal{T}_{\mathrm{measure}}$ denotes target-side geometric queries, whose results may also be included in the observation:

\begin{equation}
    o_t^{(\mathrm{full})}
    =
    \{I_t^{\mathrm{tar}}, I_t^{\mathrm{rec}}, m_t\}.
\end{equation}

Through these function calls, the agent can access properties such as bounding
boxes, transformations, dimensions, and distances. Given the interaction history $H_t^{(\mathrm{full})}$, the agent may issue a geometric query $q_t$ and obtain
\begin{equation}
    m_t
    =
    \operatorname{Measure}(G^\star, q_t).
\end{equation}
These measurements are not provided in advance; the agent must decide what to query, when to query it, and how to use the returned evidence for subsequent reconstruction.

% This setting introduces an additional form of agency beyond viewpoint selection. Unlike the preceding settings, in which target geometry must be inferred solely from visual observations, Full 3D Interaction allows explicit geometric measurements to be obtained externally through Blender MCP function calls. When a VLM can effectively exploit this functionality, reconstruction becomes easier because geometric ambiguities can be resolved directly through measurement rather than inferred from images alone.

Full 3D Interaction probes whether a VLM can combine agent-directed visual exploration, explicit geometric evidence, and reconstruction feedback to iteratively refine an executable reconstruction. Although measured quantities may be incorporated into the generated program, the reconstruction itself must remain procedurally agent-generated. The system
prompt explicitly prohibits directly copying or duplicating the ground-truth
target object into the reconstruction. The complete experimental protocol is provided in the sup. mat., including the specification of how we probe the VLMs and the complete set of Blender MCP function calls exposed in our experiments.

\section{Experiments}
\label{sec:experiments}

\begin{table*}[t]
\centering
\scriptsize
\setlength{\tabcolsep}{4pt}
\caption{\textbf{Comparison of agents across different harness settings.}
Reconstruction quality generally improves with richer target access: Full 3D
Interaction yields every agent's best geometry, while Active Visual separates
agents most sharply.
$\uparrow$ indicates higher is better, and $\downarrow$ indicates lower is better.
Bold indicates the best value within each harness, and underline indicates the second-best value.
Rankings are determined using unrounded values; displayed quality metrics are rounded to three decimal places.
Efficiency metrics and per-$\beta$ topology results are reported in the sup.\ mat.
\bingchen{added takeaway and sup.\ mat.\ pointer}}

\resizebox{0.92\textwidth}{!}{
\begin{tabular}{llccccccc}
\hline
Harness & Agent
& SIGLIP-2 $\uparrow$
& DINOv2 $\uparrow$
& DINOv3 $\uparrow$
& Chamfer $\downarrow$
& UNI3D $\uparrow$
& UNI3D I-3D $\uparrow$
& Betti Norm. L1 $\downarrow$ \\
\hline

Single-view
& \cellcolor{gray!18} Fable 5
& \cellcolor{gray!18} 0.913
& \cellcolor{gray!18} \underline{0.731}
& \cellcolor{gray!18} \textbf{0.784}
& \cellcolor{gray!18} \textbf{0.012}
& \cellcolor{gray!18} \underline{0.733}
& \cellcolor{gray!18} \underline{0.340}
& \cellcolor{gray!18} \underline{0.815} \\

& Opus 5
& 0.910
& 0.704
& 0.764
& \underline{0.015}
& 0.725
& 0.340
& 0.830 \\

& \cellcolor{gray!18} GPT-5.6 Sol
& \cellcolor{gray!18} \textbf{0.915}
& \cellcolor{gray!18} 0.714
& \cellcolor{gray!18} \underline{0.783}
& \cellcolor{gray!18} 0.016
& \cellcolor{gray!18} 0.715
& \cellcolor{gray!18} 0.332
& \cellcolor{gray!18} 1.000 \\

& Kimi K3
& 0.896
& 0.664
& 0.734
& 0.020
& 0.678
& 0.332
& 0.893 \\

& \cellcolor{gray!18} Qwen 3.8 Max
& \cellcolor{gray!18} \underline{0.913}
& \cellcolor{gray!18} \textbf{0.731}
& \cellcolor{gray!18} 0.777
& \cellcolor{gray!18} 0.017
& \cellcolor{gray!18} \textbf{0.740}
& \cellcolor{gray!18} \textbf{0.346}
& \cellcolor{gray!18} 0.835 \\

& Gemini 3.1 Pro
& 0.904
& 0.692
& 0.756
& 0.016
& 0.728
& 0.335
& \textbf{0.776} \\

& \cellcolor{gray!18} MiniMax M3
& \cellcolor{gray!18} 0.833
& \cellcolor{gray!18} 0.490
& \cellcolor{gray!18} 0.576
& \cellcolor{gray!18} 0.028
& \cellcolor{gray!18} 0.495
& \cellcolor{gray!18} 0.263
& \cellcolor{gray!18} 0.921 \\
\hline

Multi-view
& \cellcolor{gray!18} Fable 5
& \cellcolor{gray!18} \textbf{0.927}
& \cellcolor{gray!18} \underline{0.760}
& \cellcolor{gray!18} \underline{0.811}
& \cellcolor{gray!18} 0.011
& \cellcolor{gray!18} \underline{0.764}
& \cellcolor{gray!18} \textbf{0.345}
& \cellcolor{gray!18} 0.805 \\

& Opus 5
& 0.921
& 0.738
& 0.796
& \textbf{0.011}
& 0.748
& 0.343
& 0.931 \\

& \cellcolor{gray!18} GPT-5.6 Sol
& \cellcolor{gray!18} \underline{0.925}
& \cellcolor{gray!18} \textbf{0.764}
& \cellcolor{gray!18} \textbf{0.816}
& \cellcolor{gray!18} 0.015
& \cellcolor{gray!18} 0.745
& \cellcolor{gray!18} 0.326
& \cellcolor{gray!18} 1.000 \\

& Kimi K3
& 0.909
& 0.702
& 0.765
& 0.014
& 0.713
& 0.338
& 0.925 \\

& \cellcolor{gray!18} Qwen 3.8 Max
& \cellcolor{gray!18} 0.920
& \cellcolor{gray!18} 0.753
& \cellcolor{gray!18} 0.805
& \cellcolor{gray!18} \underline{0.011}
& \cellcolor{gray!18} \textbf{0.768}
& \cellcolor{gray!18} \underline{0.344}
& \cellcolor{gray!18} \underline{0.768} \\

& Gemini 3.1 Pro
& 0.908
& 0.700
& 0.774
& 0.015
& 0.753
& 0.331
& \textbf{0.736} \\

& \cellcolor{gray!18} MiniMax M3
& \cellcolor{gray!18} 0.847
& \cellcolor{gray!18} 0.522
& \cellcolor{gray!18} 0.617
& \cellcolor{gray!18} 0.022
& \cellcolor{gray!18} 0.577
& \cellcolor{gray!18} 0.299
& \cellcolor{gray!18} 0.983 \\
\hline

Active Visual
& \cellcolor{gray!18} Fable 5
& \cellcolor{gray!18} \underline{0.932}
& \cellcolor{gray!18} \underline{0.767}
& \cellcolor{gray!18} \underline{0.822}
& \cellcolor{gray!18} 0.006
& \cellcolor{gray!18} \underline{0.832}
& \cellcolor{gray!18} \textbf{0.349}
& \cellcolor{gray!18} \underline{0.663} \\

& Opus 5
& \textbf{0.939}
& \textbf{0.796}
& \textbf{0.843}
& \textbf{0.004}
& \textbf{0.871}
& \underline{0.347}
& \textbf{0.611} \\

& \cellcolor{gray!18} GPT-5.6 Sol
& \cellcolor{gray!18} 0.931
& \cellcolor{gray!18} 0.760
& \cellcolor{gray!18} 0.817
& \cellcolor{gray!18} 0.007
& \cellcolor{gray!18} 0.823
& \cellcolor{gray!18} 0.342
& \cellcolor{gray!18} 0.864 \\

& Kimi K3
& 0.925
& 0.740
& 0.802
& \underline{0.005}
& 0.819
& 0.336
& 0.685 \\

& \cellcolor{gray!18} Qwen 3.8 Max
& \cellcolor{gray!18} 0.919
& \cellcolor{gray!18} 0.729
& \cellcolor{gray!18} 0.789
& \cellcolor{gray!18} 0.006
& \cellcolor{gray!18} 0.809
& \cellcolor{gray!18} 0.331
& \cellcolor{gray!18} 0.711 \\

& Gemini 3.1 Pro
& 0.865
& 0.600
& 0.674
& 0.015
& 0.691
& 0.315
& 0.853 \\

& \cellcolor{gray!18} MiniMax M3
& \cellcolor{gray!18} 0.808
& \cellcolor{gray!18} 0.447
& \cellcolor{gray!18} 0.527
& \cellcolor{gray!18} 0.034
& \cellcolor{gray!18} 0.482
& \cellcolor{gray!18} 0.250
& \cellcolor{gray!18} 0.974 \\
\hline

Full 3D Interaction
& \cellcolor{gray!18} Fable 5
& \cellcolor{gray!18} \underline{0.949}
& \cellcolor{gray!18} \underline{0.818}
& \cellcolor{gray!18} \underline{0.859}
& \cellcolor{gray!18} \underline{0.003}
& \cellcolor{gray!18} \underline{0.884}
& \cellcolor{gray!18} \underline{0.345}
& \cellcolor{gray!18} \underline{0.557} \\

& Opus 5
& \textbf{0.957}
& \textbf{0.842}
& \textbf{0.883}
& \textbf{0.002}
& \textbf{0.927}
& \textbf{0.347}
& \textbf{0.473} \\

& \cellcolor{gray!18} GPT-5.6 Sol
& \cellcolor{gray!18} 0.939
& \cellcolor{gray!18} 0.788
& \cellcolor{gray!18} 0.841
& \cellcolor{gray!18} 0.003
& \cellcolor{gray!18} 0.879
& \cellcolor{gray!18} 0.342
& \cellcolor{gray!18} 0.677 \\

& Kimi K3
& 0.929
& 0.747
& 0.809
& 0.004
& 0.848
& 0.339
& 0.749 \\

& \cellcolor{gray!18} Qwen 3.8 Max
& \cellcolor{gray!18} 0.934
& \cellcolor{gray!18} 0.763
& \cellcolor{gray!18} 0.822
& \cellcolor{gray!18} 0.004
& \cellcolor{gray!18} 0.857
& \cellcolor{gray!18} 0.339
& \cellcolor{gray!18} 0.750 \\

& Gemini 3.1 Pro
& 0.901
& 0.694
& 0.759
& 0.008
& 0.783
& 0.322
& 0.588 \\

& \cellcolor{gray!18} MiniMax M3
& \cellcolor{gray!18} 0.857
& \cellcolor{gray!18} 0.578
& \cellcolor{gray!18} 0.651
& \cellcolor{gray!18} 0.014
& \cellcolor{gray!18} 0.679
& \cellcolor{gray!18} 0.302
& \cellcolor{gray!18} 0.821 \\
\hline

\end{tabular}
}

\label{tab:harness_comparison}
\end{table*}

\subsection{Experimental Setup}
\label{sec:experimental_setup}

\nbf{Benchmark data}
We randomly sample 100 target objects from the 212 objects in 3DCodeBench~\cite{gao2026_3dcodebench}, where each object represents a distinct category. For each target, we retain only its rendered reference images and exported GLB asset. The original procedural source program and other authoring information are never exposed to the evaluated agents. Depending on the harness, the same underlying target is presented through fixed reference renders or loaded into Blender for interactive inspection. 

\nbf{Evaluated agents}
We evaluate seven frontier VLMs spanning closed-source and open-weight models
under the four harnesses in Sec.~\ref{sec:target_access}. Each model runs in
its corresponding official agent CLI, preserving its native tool-use, context
management, and interaction mechanisms. Each instance is evaluated without cross-instance memory.

\nbf{Blender MCP environment}
Blender Model Context Protocol (MCP)~\cite{ahujasid_blender_mcp,blender_mcp_server} helps connect the agent CLIs to Blender, exposing Blender operations as callable tools.

\nbf{Quantitative metrics}
We group evaluation into 2D appearance, 3D geometry, topology metrics and interaction efficiency.
For appearance, we render four canonical views under standardized lighting and compute cosine similarity using SigLIP2~\cite{tschannen2025siglip2}, DINOv2~\cite{oquab2024dinov2}, and DINOv3~\cite{simeoni2026dinov3} features. For geometry, we report symmetric Chamfer distance~\cite{fan2017pointset} between independently centered and unit-sphere-normalized point clouds, minimized over 24 right-handed axis-aligned rotations, and Uni3D~\cite{zhou2024uni3d} cosine similarity for both 3D--3D and image--3D feature pairs. For topology, we report the median normalized $L_1$ loss over Betti numbers~\cite{hu2019topology}. API and MCP calls, reconstruction versions, output tokens, latency, and per-instance cost are reported in the sup. mat. to characterize interaction and efficiency.

\subsection{Evaluation Protocol}
\label{sec:evaluation_protocol}

We provide a detailed comparison of the four harness protocols in the
sup. mat. The protocols progress from fixed visual observations to agent-directed exploration and explicit geometric interaction.

\nbf{Single-view and Multi-view}
Both settings follow the procedural reconstruction protocol of
3DCodeBench~\cite{gao2026_3dcodebench}. Single-view provides one fixed target
view, while Multi-view provides four views separated by $90^\circ$ in yaw. Within each iteration, failed executions receive up to
three rounds of error feedback and debugging. Unlike 3DCodeBench, however, after each successful execution, we further refine the current reconstruction program using its rendered outputs together with the target renders, repeating this process for three iterations.

\nbf{Active Visual}
Active Visual uses isolated target and reconstruction Blender processes
connected through separate Blender MCP
interfaces~\cite{ahujasid_blender_mcp,blender_mcp_server}. On the target side,
the agent interacts with a \emph{constrained Blender MCP} interface, enforced
through an AST allowlist that permits only approved camera and viewport
operations. The reconstruction side retains unconstrained Blender MCP access.
This setup allows the agent to actively control viewpoint selection and zoom
while preventing direct access to target geometry.

\nbf{Full 3D Interaction}
Unlike Active Visual, which uses a constrained Blender MCP, Full 3D Interaction uses a single Blender environment connected to an \emph{unconstrained Blender MCP}~\cite{blender_mcp_server}, providing access to the full set of available MCP operations. Both the target object and the agent-generated reconstruction are maintained within the same Blender environment.

\begin{figure*}[t]
    \centering
    \includegraphics[width=\textwidth]{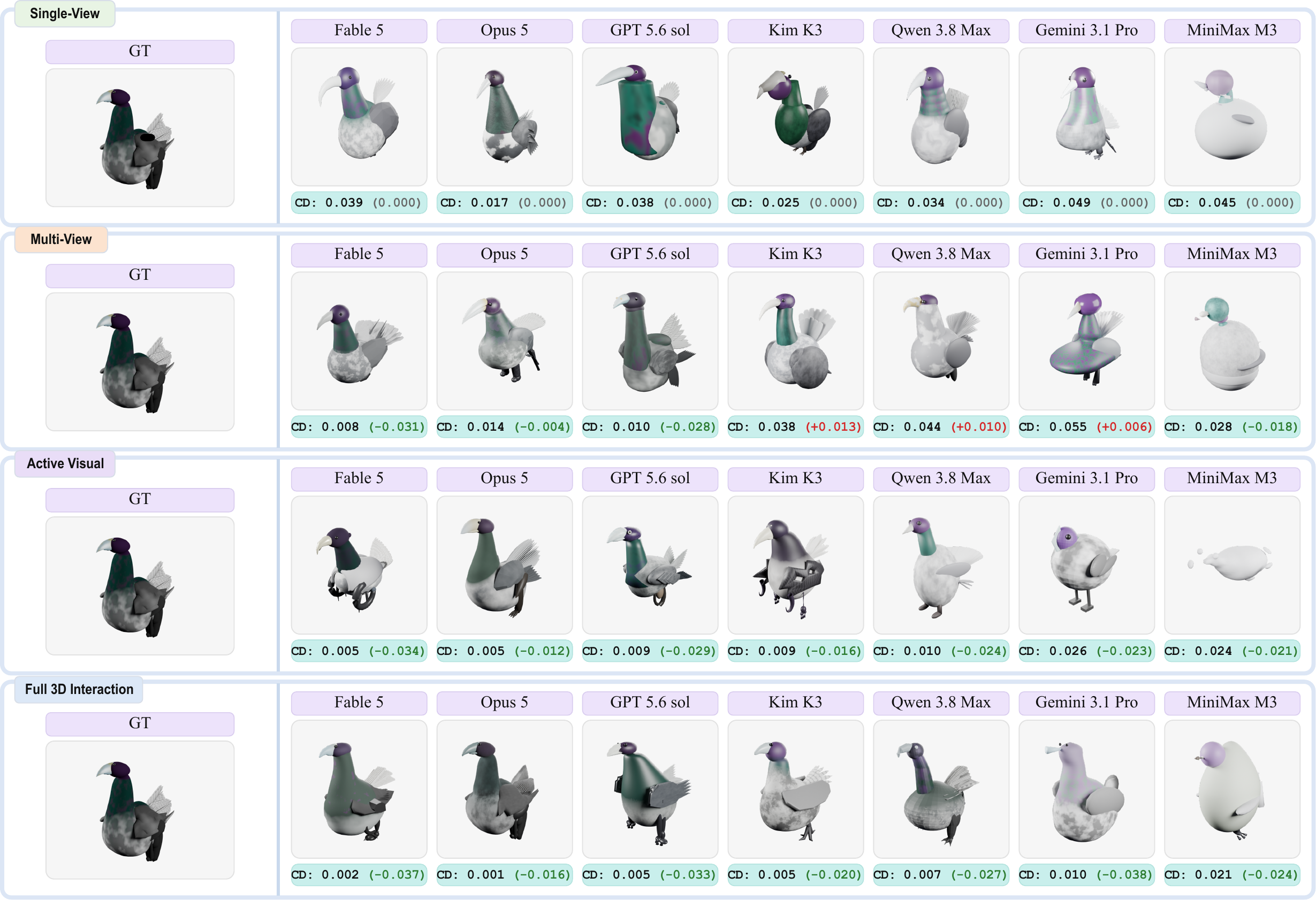}
    \caption{\textbf{Bird reconstruction from seven frontier VLM agents across four harness settings.}
    Numbers below each
    reconstruction report the Chamfer distance (CD), with the difference from
    the corresponding Single-view result in parentheses (green: lower, red:
    higher). For this example, Active Visual and Full 3D Interaction enable Fable 5, Opus 5, GPT-5.6 Sol, Kimi K3, and Qwen 3.8 Max to better reconstruct the target's body proportions and rear wing structure, while Gemini 3.1 Pro remains simplified and MiniMax M3 largely fails to recover the bird's articulated structure even with full target access.}
    \label{fig:bird_visualization}
\end{figure*}

%%%%%%%%%
\subsection{Main Results}
\label{sec:main_results}

As shown in Table~\ref{tab:harness_comparison}, 2D appearance, 3D geometry,
and topology exhibit broadly similar trends as target access becomes richer.

\nbf{Effect of target access}
Within each agent, Multi-view generally improves appearance and geometry over Single-view. Active Visual further separates agents in their ability to exploit agent-directed observations: Opus 5 gains most consistently, with Fable 5 and Kimi K3 also benefiting strongly; GPT-5.6 Sol improves mainly in geometry, while Qwen 3.8 Max shows mixed changes. In contrast, Gemini 3.1 Pro and MiniMax M3 regress on most metrics, indicating that richer visual access helps only when agents can effectively acquire and integrate the resulting evidence.

Full 3D Interaction yields the strongest overall results for every agent. Figure~\ref{fig:bird_visualization} shows the same qualitative progression: fixed-view reconstructions often capture only coarse shape and visible parts, whereas Active Visual and especially Full 3D Interaction recover more faithful part sizes, structures, and geometry. Additional qualitative results for all seven agents and targets are provided in the sup. mat.

\nbf{Cross-agent comparison}
Performance is relatively clustered under fixed-view settings, except for MiniMax M3, whereas interactive settings reveal substantially larger differences. Opus 5 is strongest overall under both Active Visual and Full 3D Interaction; Fable 5, GPT-5.6 Sol, Kimi K3, and Qwen 3.8 Max form a competitive second group, while Gemini 3.1 Pro and MiniMax M3 remain weaker, particularly under Active Visual.

From Multi-view to Active Visual, Uni3D changes by $+16.4\%$ for Opus 5, $+14.9\%$ for Kimi K3, $+10.5\%$ for GPT-5.6 Sol, $+8.9\%$ for Fable 5, and $+5.3\%$ for Qwen 3.8 Max, but by $-8.2\%$ for Gemini 3.1 Pro and $-16.5\%$ for MiniMax M3. Consequently, the weakest-to-strongest Uni3D gap reaches $44.7\%$ under Active Visual, versus $24.9\%$ for Multi-view, $33.1\%$ for Single-view, and $26.8\%$ for Full 3D Interaction; the worst-to-best Chamfer ratio likewise grows from roughly $2\times$ in fixed-view settings to $8.5\times$. Active Visual is therefore the sharpest discriminator: additional access helps some agents but hurts others, exposing their ability to convert tool-mediated visual evidence into effective reconstruction updates. Coding and tool-use competence may contribute to this gap, as interactive harnesses additionally require correct tool calls and Blender control.

We further compare our results with those reported in 3DCodeBench~\cite{gao2026_3dcodebench} in the sup. mat.

\subsection{Analysis}
\label{sec:analysis}

\label{sec:analysis_pose}
\nbf{Geometric Grounding Improves Multi-view 3D Reconstruction}
To examine whether additional geometric grounding contributes to the gains
beyond  Multi-view input, we conduct this
ablation on a subset of 10 benchmark instances using GPT-5.6 Sol. The
\emph{Baseline} follows the standard Multi-view setting with four target views
and three refinement iterations. \emph{Grid} uses the same four target views rendered in the Blender viewport with the world-scale grid enabled, providing an explicit metric-scale reference that is absent from the standard Multi-view renders. \emph{Pose} uses the Multi-view renders together with the corresponding camera
parameters, while \emph{Grid + pose} combines both sources of geometric
reference. Table~\ref{tab:ablation_grid_pose} shows that camera pose provides the most
significant gains in appearance and geometry, whereas grid cues yield smaller improvements. Combining grid and pose does not further
improve appearance or geometry and produces the largest topological error.
Overall, remarkably, explicit camera pose information is more useful than viewport grid cues for
Multi-view reconstruction, while the two signals provide limited additive
benefit.

\label{sec:analysis_active_observation}
\nbf{Active Visual Performance Depends on Tool Calling and Observation Quality}
We disentangle Active Visual failure modes on Gemini 3.1 Pro and MiniMax M3 over 10 instances. \emph{Multi-view} and \emph{Active Visual} use the standard settings. \emph{Normalized Active Visual} preserves the agent's viewpoint-selection behavior while normalizing the viewport before each screenshot to control observation quality. \emph{Fable 5 Inspection} instead provides inspection renders collected by Fable 5, testing whether agents better exploit observations from a stronger inspection policy. Example screenshots are provided in the sup. mat.

Table~\ref{tab:observation_ablation} reveals distinct failure sources. Normalized Active Visual substantially improves MiniMax M3 and even better than Multi-view, showing that poor viewport states can degrade observations during tool interaction. Fable 5's inspection trajectory further improves Gemini 3.1 Pro, recovering its Multi-view performance on several metrics. Overall, the degradation from Multi-view to Active Visual of both agents is largely driven by observation acquisition and quality, where Gemini 3.1 Pro generally follows a valid interaction loop but performs little inspection and tends to terminate early; further analysis is provided in the sup. mat.

\begin{table}[t]
\centering
\caption{\textbf{Pose and grid cues in Multi-view reconstruction} (10
instances, GPT-5.6 Sol). Camera poses give the most consistent gains; grid
cues help less reliably, and combining both is not additive. Full metrics in
the sup.\ mat.}
\label{tab:ablation_grid_pose}
\resizebox{\columnwidth}{!}{%
\begin{tabular}{lcccc}
\toprule
Setting
& SIGLIP-2 $\uparrow$
& Chamfer $\downarrow$
& UNI3D $\uparrow$
& Betti Norm. L1 $\downarrow$ \\
\midrule
Baseline
& 0.9254
& 0.0199
& 0.7038
& 0.7567 \\

+ Grid
& 0.9321
& 0.0172
& 0.8351
& \textbf{0.7487} \\

+ Pose
& \textbf{0.9419}
& \textbf{0.0077}
& \textbf{0.8387}
& 0.8071 \\

+ Grid + Pose
& 0.9391
& 0.0110
& 0.8098
& 0.9104 \\
\bottomrule
\end{tabular}%
}
\end{table}

% \vspace{-10pt}
\begin{table}[t]
\centering
\caption{\textbf{Observation-interface ablation under Active Visual} (10
instances). Normalized viewport screenshots and Fable 5's inspection views
improve both weaker agents over the Active Visual baseline on most metrics.
Full metrics in the sup.\ mat.}
\label{tab:observation_ablation}
\scriptsize
\setlength{\tabcolsep}{2.0pt}
\renewcommand{\arraystretch}{0.95}

\resizebox{\columnwidth}{!}{%
\begin{tabular}{llcccc}
\toprule
\textbf{Model}
& Setting
& SIGLIP-2 $\uparrow$
& Chamfer $\downarrow$
& UNI3D $\uparrow$
& Betti Norm. L1 $\downarrow$ \\
\midrule

\multirow{4}{*}{Gemini 3.1 Pro}
& Multi-view
& 0.9189
& \textbf{0.0165}
& \textbf{0.7609}
& 0.6837 \\

& Active Visual
& 0.8452
& 0.0234
& 0.6776
& \textbf{0.6116} \\

& Normalized Active Visual
& 0.8710
& 0.0186
& 0.7321
& 0.9065 \\

& Fable 5 Inspection
& \textbf{0.9232}
& 0.0191
& 0.7605
& 0.6793 \\

\midrule

\multirow{4}{*}{MiniMax M3}
& Multi-view
& 0.8230
& 0.0262
& \textbf{0.6200}
& 0.9696 \\

& Active Visual
& 0.7583
& 0.0798
& 0.3722
& 0.9863 \\

& Normalized Active Visual
& \textbf{0.8315}
& \textbf{0.0240}
& 0.6148
& \textbf{0.8326} \\

& Fable 5 Inspection
& 0.8279
& 0.0414
& 0.4899
& 0.9541 \\

\bottomrule
\end{tabular}%
}
\end{table}

\label{sec:analysis_measurement} 
\nbf{Explicit Geometric Measurements Provide Reconstruction Gains}
Table~\ref{tab:ablation_measurement} evaluates explicit geometric measurements added to Multi-view input on 10 instances with Opus 5. Global measurements include the target bounding box, center, and dimensions, while parts measurements provide anonymous part centers, dimensions, and component or repetition statistics extracted from the ground-truth mesh using a GPT-5.6 Sol-authored measurement program. Explicit measurements consistently improve almost all the metrics. Parts measurement achieves the best Uni3D and topological error, while combining global and part-level cues gives the lowest Chamfer distance. Global Measurement, hoewever, achieve the highest SIGLIP-2.  Overall, global and part measurements provide geometric information, with the clearest gains in geometry and topology.

\begin{table}[t]
\centering
\caption{\textbf{Explicit geometric measurements added to Multi-view input}
(10 instances, Opus 5). Adding explicit geometric measurements improves reconstruction quality over image-only input. Full metrics in the sup.\ mat.}
% with global and part-level measurements providing complementary benefits. Their combination achieves the best overall geometric accuracy. Full metrics are reported in the supplementary material.}

\label{tab:ablation_measurement}
\resizebox{\columnwidth}{!}{%
\scriptsize
\setlength{\tabcolsep}{2.2pt}
\begin{tabular}{lcccc}
\toprule
Setting
& SIGLIP-2 $\uparrow$
& Chamfer $\downarrow$
& UNI3D $\uparrow$
& Betti Norm. L1 $\downarrow$ \\
\midrule
Imgs
& 0.9173
& 0.0150
& 0.8133
& 0.9050 \\

Imgs + Global
& \textbf{0.9309}
& 0.0103
& 0.8088
& 0.7093 \\

Imgs + Parts
& 0.9256
& 0.0099
& \textbf{0.8557}
& \textbf{0.4186} \\

Imgs + Global + Parts
& 0.9307
& \textbf{0.0084}
& 0.8296
& 0.5783 \\
\bottomrule
\end{tabular}
}
\end{table}

\begin{figure}[t]
    \centering
    \includegraphics[width=\linewidth]{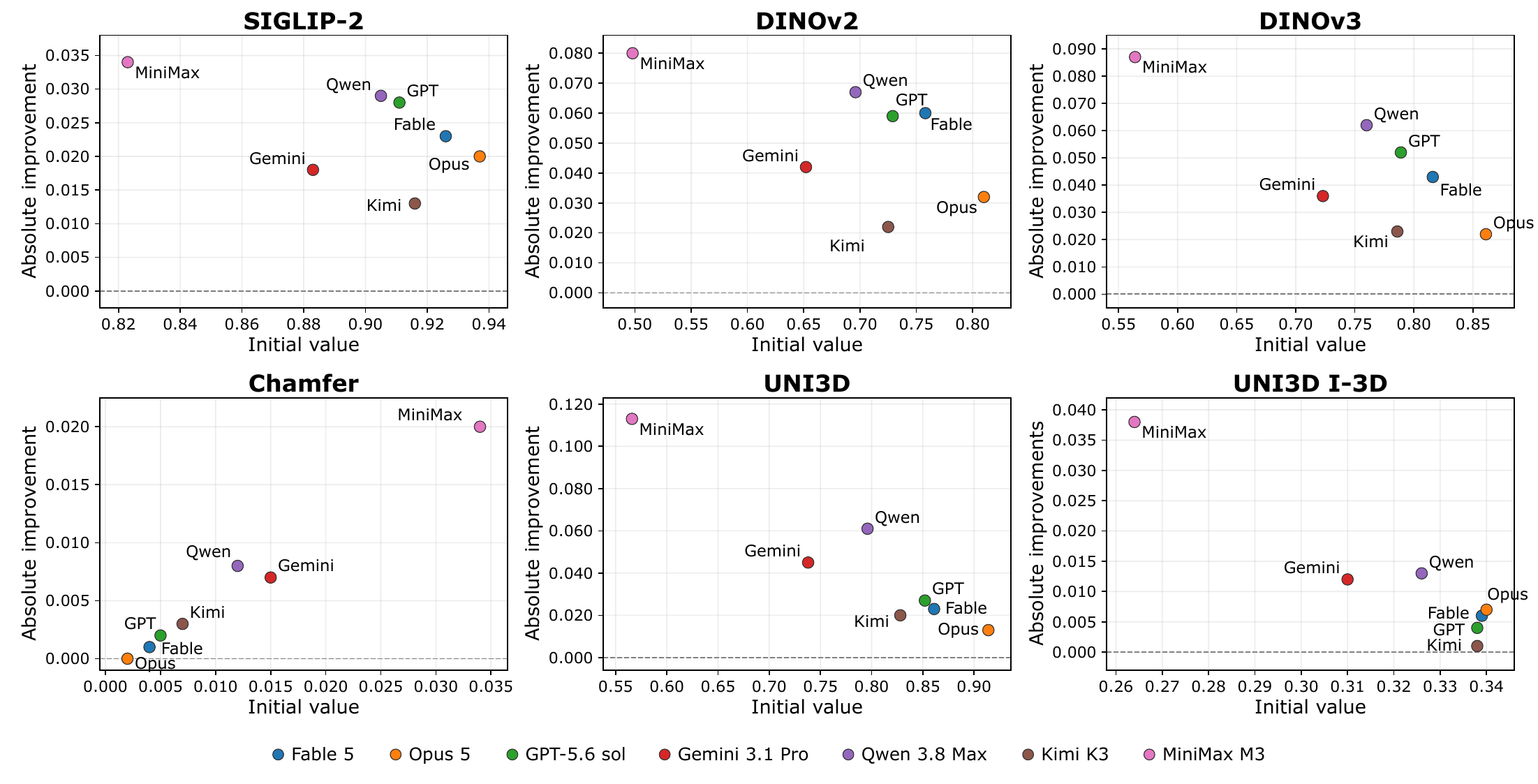}
    \caption{\textbf{Initial reconstruction quality and final refinement gain under Full 3D Interaction.} Each point represents one agent, with its initial reconstruction score on the x-axis and the absolute improvement on the y-axis. }
    % Weaker initializers benefit most from refinement (MiniMax M3), while the strongest (Opus 5, Fable 5) gain least. Qwen 3.8 Max uniquely combines strong initialization with large gains.}
    \label{fig:self-correction}
    \vspace{-20pt}
\end{figure}

\label{sec:analysis_refinement} 
\nbf{Initial Reconstruction and Interactive Refinement Are Distinct Agent Capabilities}
To analyze agents' initial reconstruction and self-correction abilities, we
separately evaluate the first executable reconstruction and the final result
after multi-turn refinement under Full 3D Interaction.
Figure~\ref{fig:self-correction} compares initial reconstruction quality with
the relative improvement achieved through refinement. Opus 5 exhibits the
strongest initialization overall, followed by Fable 5, while GPT-5.6 Sol and
Kimi K3 also begin from relatively strong reconstructions. MiniMax M3, in
contrast, has the weakest initialization across most metrics.

Refinement ability follows a different pattern. MiniMax M3 achieves the largest gains on most metrics despite its weak initialization, while Qwen 3.8 Max shows particularly strong self-correction, including the largest reduction in Chamfer distance. Opus 5 and Fable 5 improve less from their already strong initial reconstructions, whereas GPT-5.6 Sol, Gemini 3.1 Pro, and Kimi K3 show more moderate refinement. These results suggest that initialization quality and self-correction capture distinct aspects of agent capability. We further visualize reconstruction trajectories across intermediate versions in the sup. mat.

\section{Conclusion and Future Work}

We introduced \textbf{3DHarnessBench} to evaluate agentic 3D-to-code capabilities of frontier VLMs under progressively richer target access. Multi-view generally improves over Single-view, while Active Visual shows that richer observations help only when effectively acquired and integrated; Full 3D Interaction performs best overall. Ablations further show gains from geometric grounding and explicit measurements, and reveal initial reconstruction and interactive self-correction as distinct capabilities. Future work will extend to more complex scenes involving multi-object placement and affordance-based constraints, and will explore fine-tuning on benchmark agentic traces to improve the performance of smaller models.

% We introduced \textbf{3DHarnessBench}, a benchmark for probing agentic 3D-to-code capabilities of frontier VLMs under progressively richer target access. Across seven agents, Multi-view generally improves reconstruction over Single-view, while Active Visual reveals that additional observations are useful only when agents can effectively acquire and integrate them. Full 3D Interaction achieves the strongest overall performance, and our ablations show
% that geometric grounding and explicit measurements provide clear gains in 3D geometry and topology, for VLMs that can exploit  them. We further find that initial reconstruction and
% interactive self-correction are distinct agent capabilities. 

% One limitation is that our framework currently focuses on individual objects, whereas future work can also consider complex scenes with multi-object placement and affordance-based constraints. Additionally, it will be interesting to fine-tune existing models on agentic traces collected in our benchmark to potentially improve performance of smaller models.

%Overall, the
% results show that strong agentic 3D reconstruction requires not only richer
% evidence, but also the ability to acquire, interpret, and translate that
% evidence into an executable 3D program.
\section{Acknowledgments}

Parts of this work were supported by the ERC Consolidator Grant 101087347 (VEGA), Institut Carnot TSN, as well as gifts from Ansys Inc., and Adobe Research.

% \clearpage
\newpage
{
    \small
    \bibliographystyle{ieeenat_fullname}
    \bibliography{main}
}

\clearpage
\setcounter{page}{1}
\maketitlesupplementary

\section{Overview}
\label{sec:supp_overview}
This supplementary material provides additional details and analyses
supporting the main paper, organized as follows.
\begin{itemize}
    \item \textbf{Sec.~\ref{sec:supp_protocol} -- Evaluation Protocol}:
    supplementing the benchmark formulation and harness settings in
    Secs.~3.1--3.2 and the experimental protocol in Secs.~4.1--4.2 of the
    main paper, we provide the per-harness protocol table (including the
    3DCodeBench~\cite{gao2026_3dcodebench} protocol for reference), the complete Blender MCP interface
    specification, and the experimental settings and metric definitions behind
    all reported results.

    \item \textbf{Sec.~\ref{sec:supp_results} -- Main Results}:
    supplementing Sec.~4.3 and Table~1 of the main paper, we report the
    per-Betti-number topology errors and interaction-efficiency statistics for
    all agents and harnesses. We also provide the comparison with 3DCodeBench
    referenced in Sec.~4.3 and a complementary VLM-based perceptual evaluation.

    \item \textbf{Sec.~\ref{sec:supp_ablation_results} -- Ablations and
    Analyses}:
    extending the analyses in Sec.~4.4 of the main paper, we report the complete
    metric sets for the camera-pose, grid, observation-interface, and explicit
    geometric-measurement ablations. We further analyze the Active Visual
    failure modes through inspection tool-calling statistics and trajectories,
    and provide the intermediate reconstruction trajectories referenced in the
    initialization-versus-refinement analysis.

    \item \textbf{Sec.~\ref{sec:supp_visualization} -- Additional Qualitative
    Results}:
    following the additional qualitative results referenced in Sec.~4.3 and Figure.~3 of the
    main paper, we provide further reconstruction comparisons across harness
    settings and agents, including additional examples under Full 3D
    Interaction.

    \item \textbf{Sec.~\ref{sec:supp_prompts} -- Prompts and Agent Skill}:
    providing the complete protocol details promised in Sec.~3.2.3 of the main
    paper, we reproduce the exact prompts used by the fixed-view harnesses and
    the task message and Agent Skill, together with its reference documents,
    used by the interactive harnesses. We additionally provide the VLM-judge
    prompt used for the supplementary perceptual evaluation.

    \item \textbf{Sec.~\ref{sec:supp_code} -- Example of Agent-Generated Code}:
    we provide
    a complete Blender Python reconstruction program generated by an evaluated
    agent under Full 3D Interaction.

\end{itemize}

\section{Evaluation Protocol}
\label{sec:supp_protocol}

\subsection{Harness Protocols}
\label{sec:supp_eval_protocol}

\begin{table*}[t]
    \centering
    \small
    \setlength{\tabcolsep}{6pt}
    \caption{\textbf{Harness protocols of 3DHarnessBench, with the 3DCodeBench
    image-to-3D protocol for reference.} Target-side access expands from fixed
    renders to agent-controlled viewing and geometric queries, while all
    harnesses share the same executable reconstruction objective; the
    interactive settings replace fixed refinement rounds with open-ended
    interaction. Our fixed-view settings inherit the 3DCodeBench interface but
    add three render-and-refine iterations.}
    \label{tab:evaluation_protocol}
    \resizebox{\textwidth}{!}{%
    \begin{tabular}{lcccc}
        \toprule
        Setting
        & Target evidence
        & Target-side access
        & Reconstruction-side access
        & Refinement \\
        \midrule

        3DCodeBench~\cite{gao2026_3dcodebench} (image-to-3D)
        & 4 fixed views
        & Images
        & Program execution + render
        & Execution debugging only \\
        \midrule

        Single-view
        & 1 fixed view
        & Image
        & Program execution + render
        & 3 refine iterations \\

        Multi-view
        & 4 fixed views
        & Images
        & Program execution + render
        & 3 refine iterations \\

        Active Visual
        & Agent-selected views
        & Constrained Blender MCP (viewport + screenshots)
        & Unconstrained Blender MCP
        & Interactive \\

        Full 3D Interaction
        & Views + geometric queries
        & Unconstrained Blender MCP
        & Unconstrained Blender MCP
        & Interactive \\

        \bottomrule
    \end{tabular}%
    }
\end{table*}

Table~\ref{tab:evaluation_protocol} summarizes the four harness protocols,
which progress from fixed visual observations to agent-directed exploration
and explicit geometric interaction, alongside the fixed-view image-to-3D
protocol of 3DCodeBench~\cite{gao2026_3dcodebench} from which our fixed-view
settings are derived. Single-view and Multi-view adopt the initial
reconstruction prompt of 3DCodeBench and the editing prompt of
BlenderGym~\cite{gu2025blendergym}; Active Visual and Full 3D Interaction use
a shared Agent Skill. All prompts are reproduced verbatim in
Sec.~\ref{sec:supp_prompts}.

\subsection{Blender MCP Interface}
\label{sec:supp_mcp}

We use the official Blender Lab MCP server~\cite{blender_mcp_server} (release v1.0.0), which connects an MCP-compatible agent to a running Blender instance over standard I/O and exposes Blender functionality, including the execution of Blender Python code, scene and object inspection, viewport screenshots, navigation, rendering, and access to Blender documentation. Table~\ref{tab:mcp_tools} summarizes the MCP tools exposed by the server in our evaluation.

Specifically, in the Active Visual setting, the target-side Blender instance is
exposed through a separate, restricted MCP server~\cite{ahujasid_blender_mcp}
that provides only two functions: a viewport-screenshot tool
(\texttt{get\_viewport\_screenshot}) and
\texttt{execute\_blender\_code}. For the latter, the server validates every
submitted script against an abstract-syntax-tree (AST) allow-list before
forwarding it to Blender. This restricts execution to approved viewport and
camera operations, including view direction, zoom, shading mode, and overlays;
all other Blender Python operations are rejected.

This restricted target-side server is derived from the third-party
\texttt{blender-mcp} implementation~\cite{ahujasid_blender_mcp} (v1.8.0),
while the Blender add-on used in our setup is v1.2.0. On the reconstruction
side of Active Visual, as well as under Full 3D Interaction, the agent has
access to the full set of MCP function calls listed in
Table~\ref{tab:mcp_tools}.

\begin{table*}[t]
\centering
\small
\setlength{\tabcolsep}{4pt}
\renewcommand{\arraystretch}{0.95}
\caption{\textbf{Blender MCP function calls used in our evaluation.} In Active Visual, target-side code execution is restricted to viewport and camera control, whereas Full 3D Interaction provides access to the complete set of Blender MCP functions.}
\label{tab:mcp_tools}

\begin{tabular}{@{}llcc@{}}
\toprule
Category
& MCP tool family
& \shortstack{Active Visual\\Target-side}
& \shortstack{Full 3D Interaction / Active Visual\\Reconstruction-side} \\
\midrule

Code execution
& \texttt{execute\_blender\_code}
& Restricted$^\dagger$
& \checkmark \\

Scene inspection
& \texttt{get\_*summary} (2)
& --
& \checkmark \\

Blend-file inspection
& \texttt{get\_blendfile\_summary\_*} (5)
& --
& \checkmark \\

Visual observation
& \texttt{get\_screenshot\_*\_as\_image} (2)
& \checkmark$^\ddagger$
& \checkmark \\

UI-state inspection
& \texttt{get\_screenshot\_*\_as\_json} (1)
& --
& \checkmark \\

Navigation
& \texttt{jump\_to\_*} (4)
& --
& \checkmark \\

Rendering
& \texttt{render\_*\_to\_path} (2)
& --
& \checkmark \\

Documentation
& \texttt{*\_docs} (3)
& --
& \checkmark \\

\bottomrule
\end{tabular}

\vspace{2pt}
\footnotesize
$^\dagger$ code execution is restricted to viewport and camera control (AST allow-list). $^\ddagger$ exposed on the target side as the single \texttt{get\_viewport\_screenshot} tool of the restricted server (Sec.~\ref{sec:supp_mcp}).
\end{table*}

\subsection{Experimental Settings}
\label{sec:supp_settings}

For reproducibility, we summarize the settings shared by all reported
results.

\paragraph{Benchmark data.}
We use 100 target objects randomly sampled from the 212 objects of
3DCodeBench~\cite{gao2026_3dcodebench}, one per category. For each target we
keep only the rendered reference images and the exported GLB asset; the
original procedural source program is never exposed to the agents. The same
100 targets are used in all four harnesses, and all ablations
(Sec.~\ref{sec:supp_ablation_results}) use the same fixed 10-instance subset:
AgaveMonocot, Auger, Bird, BushBaseCoral, CantileverStaircase, Crustacean,
FlowerPlant, FruitDurian, WheatEarMonocot, and WheatMonocot.

\paragraph{Agents.}
We evaluate seven frontier VLMs, each driven through an agent CLI with the
CLI's default sampling settings and its reasoning-effort control set to
\emph{high}: Fable~5 and Opus~5 through Claude Code 2.1.226--2.1.228; GPT-5.6 Sol through Codex
CLI 0.147.0; Gemini 3.1 Pro through the Antigravity
CLI \texttt{agy} 1.0.16; Kimi K3
through Kimi Code CLI 0.34.0; Qwen 3.8 Max through Qwen Code
0.21.0; and MiniMax M3 through the MiniMax CLI \texttt{mmx} 1.0.19. The same agent stack is used in Active Visual and Full 3D Interaction settings. Every instance is evaluated in a fresh session without
cross-instance memory.

\paragraph{Fixed-view harnesses.}
Single-view provides one target render; Multi-view provides four. Both use
the textured turntable renders of 3DCodeBench at azimuths of $45^\circ$,
$135^\circ$, $225^\circ$, and $315^\circ$, and Single-view uses the
$45^\circ$ view. Each instance runs one initial generation followed by three
render-and-refine iterations, each allowing up to three rounds of
execution-error feedback.
Generated scripts are executed headlessly with Blender 5.1.2. The renders
fed back to the agent are produced at $512\times512$ from the same canonical
views.

\paragraph{Interactive harnesses.}
Active Visual runs the target and the reconstruction in two separate Blender
processes with the target-side restriction described in
Sec.~\ref{sec:supp_mcp}; Full 3D Interaction runs both in a single Blender
process with the unrestricted tool set. Both use Blender 5.1.2 and the
official Blender MCP server v1.0.0 on the reconstruction side. There is no cap
on the number of turns or tool calls: the interaction ends when the agent
reports completion. Each agent turn is bounded by a 3600-second harness
timeout; when it fires, the Blender state is checkpointed and the same CLI
session is resumed in place, up to five times per timeout episode.

\paragraph{Metrics.}
For appearance, target and reconstruction are each rendered from the four
canonical turntable views (azimuths $45^\circ$, $135^\circ$, $225^\circ$, and
$315^\circ$; camera distance $1.8\times$ and height $0.6\times$ the object
extent) under identical three-point lighting at $512\times512$ resolution with
Blender 5.1.2. The object extent is defined as the maximum side length of its
axis-aligned bounding box. We use three white Area Lights, all oriented toward
the object center: a key light at
$(1.62,-1.26,1.02)\times$ the object extent with size
$0.9\times$ the object extent and base power $1200$\,W; a fill light at
$(-1.08,-0.72,0.60)\times$ the object extent with size
$1.2\times$ the object extent and base power $400$\,W; and a rim light at
$(0,1.62,0.78)\times$ the object extent with size
$0.7\times$ the object extent and base power $600$\,W. The actual power of
each light is scaled according to
\[
P_{\mathrm{light}}
=
P_{\mathrm{base}} \cdot
\left(\frac{\text{object extent}}{2.5}\right)^2
,
\]
where $P_{\mathrm{base}}$ is the corresponding base power listed above.
We compute the cosine similarity between
SigLIP2~\cite{tschannen2025siglip2} (\texttt{so400m-patch16-naflex}),
DINOv2~\cite{oquab2024dinov2} (ViT-L/14), and
DINOv3~\cite{simeoni2026dinov3} (ViT-L/16) image embeddings. Because an agent
may place its reconstruction in a different canonical orientation, the four
rendered views are matched to the four reference views by the best one-to-one
assignment over the $4\times4$ similarity matrix, and the per-instance score is
the mean over the matched pairs. For geometry, we sample 8192 surface points
with a fixed random seed from the target and reconstructed meshes, independently
center each point cloud on its centroid and scale it to the unit sphere, and
report the symmetric Chamfer distance, defined as the mean squared
nearest-neighbor distance in both directions, minimized over the 24
right-handed axis-aligned rotations of the reconstruction. We additionally
report Uni3D-Giant~\cite{zhou2024uni3d} cosine similarity for the 3D--3D pair
of point clouds and for the image--3D pair formed by the $45^\circ$ target
render and the reconstructed point cloud. For topology, we compute the
$\mathbb{Z}_2$ Betti numbers $(\beta_0,\beta_1,\beta_2)$ of each triangle mesh
from its simplicial boundary maps after merging coincident vertices, and report
the normalized $L_1$ error
$\sum_i|\beta_i^{\mathrm{gt}}-\beta_i^{\mathrm{rec}}|/
\sum_i\beta_i^{\mathrm{gt}}$
as the median over instances; all other metrics are averaged over instances.
An instance whose final program yields no loadable mesh receives zero
similarity and $1.5\times$ the worst valid Chamfer distance of the same run.
For interaction efficiency, API Calls denotes calls from the agent CLI to the
underlying model, MCP API Calls denotes Blender MCP function calls, and Versions
counts all reconstruction versions produced by the agent, including
non-executable ones; we additionally report output tokens, monetary cost, and
end-to-end latency.

\begin{table*}[t]
\centering
\small
\caption{\textbf{Topology and interaction-efficiency comparison across harness settings.}
Cost is the session cost reported by the agent CLI where available; $^{*}$ marks costs estimated from the reported token counts. ``--'' denotes values that the CLI does not report or that do not apply (MCP calls in the fixed-view harnesses).
}

\resizebox{\textwidth}{!}{%
\begin{tabular}{llccccccccc}
\toprule
Harness & Agent
& \multicolumn{3}{c}{Betti Number Normalized $L_1$ $\downarrow$}
& API Calls 
& MCP API Calls 
& Versions 
& Output Tokens $\downarrow$
& Cost $\downarrow$
& Latency (s) $\downarrow$ \\
\cmidrule(lr){3-5}
& & $\beta_0$ & $\beta_1$ & $\beta_2$
& & & & & & \\
\midrule

Single-view
& \cellcolor{gray!18} Fable 5
& \cellcolor{gray!18} 0.105
& \cellcolor{gray!18} \textbf{0.021}
& \cellcolor{gray!18} 0.255
& \cellcolor{gray!18} 11
& \cellcolor{gray!18} --
& \cellcolor{gray!18} 4.14
& \cellcolor{gray!18} 51,846
& \cellcolor{gray!18} 2.00
& \cellcolor{gray!18} 382 \\

& Opus 5
& 0.140
& 0.074
& 0.250
& 12
& --
& 4.32
& 101,440
& 2.04
& 780 \\

& \cellcolor{gray!18} GPT-5.6 Sol
& \cellcolor{gray!18} 0.333
& \cellcolor{gray!18} 0.097
& \cellcolor{gray!18} 0.431
& \cellcolor{gray!18} 4
& \cellcolor{gray!18} --
& \cellcolor{gray!18} 4
& \cellcolor{gray!18} 30,856
& \cellcolor{gray!18} 1.26$^{*}$
& \cellcolor{gray!18} 664 \\

& Kimi K3
& 0.233
& 0.077
& 0.318
& 15
& --
& 4.39
& 28,199
& --
& 928 \\

& \cellcolor{gray!18} Qwen 3.8 Max
& \cellcolor{gray!18} \textbf{0.098}
& \cellcolor{gray!18} 0.042
& \cellcolor{gray!18} 0.262
& \cellcolor{gray!18} 4
& \cellcolor{gray!18} --
& \cellcolor{gray!18} 4.41
& \cellcolor{gray!18} 39,959
& \cellcolor{gray!18} --
& \cellcolor{gray!18} 758 \\

& Gemini 3.1 Pro
& 0.125
& 0.060
& \textbf{0.248}
& 4
& --
& 4.57
& --
& --
& 429 \\

& \cellcolor{gray!18} MiniMax M3
& \cellcolor{gray!18} 0.198
& \cellcolor{gray!18} 0.225
& \cellcolor{gray!18} 0.360
& \cellcolor{gray!18} 6
& \cellcolor{gray!18} --
& \cellcolor{gray!18} 6.65
& \cellcolor{gray!18} 16,640
& \cellcolor{gray!18} --
& \cellcolor{gray!18} 217 \\

\midrule

Multi-view
& \cellcolor{gray!18} Fable 5
& \cellcolor{gray!18} 0.108
& \cellcolor{gray!18} 0.061
& \cellcolor{gray!18} 0.241
& \cellcolor{gray!18} 33
& \cellcolor{gray!18} --
& \cellcolor{gray!18} 4.11
& \cellcolor{gray!18} 68,625
& \cellcolor{gray!18} 2.54
& \cellcolor{gray!18} 428 \\

& Opus 5
& 0.167
& 0.060
& 0.338
& 34
& --
& 4.19
& 130,053
& 2.55
& 946 \\

& \cellcolor{gray!18} GPT-5.6 Sol
& \cellcolor{gray!18} 0.335
& \cellcolor{gray!18} 0.073
& \cellcolor{gray!18} 0.387
& \cellcolor{gray!18} 4
& \cellcolor{gray!18} --
& \cellcolor{gray!18} 4
& \cellcolor{gray!18} 34,308
& \cellcolor{gray!18} 1.47$^{*}$
& \cellcolor{gray!18} 744 \\

& Kimi K3
& 0.203
& 0.066
& 0.282
& 18
& --
& 4.70
& 30,410
& --
& 1065 \\

& \cellcolor{gray!18} Qwen 3.8 Max
& \cellcolor{gray!18} \textbf{0.102}
& \cellcolor{gray!18} \textbf{0.040}
& \cellcolor{gray!18} \textbf{0.204}
& \cellcolor{gray!18} 4
& \cellcolor{gray!18} --
& \cellcolor{gray!18} 4.24
& \cellcolor{gray!18} 41,173
& \cellcolor{gray!18} --
& \cellcolor{gray!18} 935 \\

& Gemini 3.1 Pro
& 0.128
& 0.070
& 0.231
& 4
& --
& 4.56
& --
& --
& 587 \\

& \cellcolor{gray!18} MiniMax M3
& \cellcolor{gray!18} 0.228
& \cellcolor{gray!18} 0.158
& \cellcolor{gray!18} 0.361
& \cellcolor{gray!18} 6
& \cellcolor{gray!18} --
& \cellcolor{gray!18} 6.81
& \cellcolor{gray!18} 18,652
& \cellcolor{gray!18} --
& \cellcolor{gray!18} 356 \\

\midrule

Active Visual
& \cellcolor{gray!18} Fable 5
& \cellcolor{gray!18} 0.146
& \cellcolor{gray!18} 0.021
& \cellcolor{gray!18} \textbf{0.098}
& \cellcolor{gray!18} 58
& \cellcolor{gray!18} 51
& \cellcolor{gray!18} 7.21
& \cellcolor{gray!18} 153,361
& \cellcolor{gray!18} 7.93
& \cellcolor{gray!18} 860 \\

& Opus 5
& 0.095
& 0.018
& 0.155
& 116
& 87
& 17.20
& 328,147
& 10.74
& 1790 \\

& \cellcolor{gray!18} GPT-5.6 Sol
& \cellcolor{gray!18} 0.158
& \cellcolor{gray!18} 0.080
& \cellcolor{gray!18} 0.272
& \cellcolor{gray!18} 55
& \cellcolor{gray!18} 69
& \cellcolor{gray!18} 15.09
& \cellcolor{gray!18} 45,004
& \cellcolor{gray!18} 3.24$^{*}$
& \cellcolor{gray!18} 939 \\

& Kimi K3
& 0.089
& \textbf{0.009}
& 0.163
& 114
& 41
& 12.61
& 53,444
& --
& 2370 \\

& \cellcolor{gray!18} Qwen 3.8 Max
& \cellcolor{gray!18} 0.125
& \cellcolor{gray!18} 0.041
& \cellcolor{gray!18} 0.167
& \cellcolor{gray!18} 393
& \cellcolor{gray!18} 55
& \cellcolor{gray!18} 11.63
& \cellcolor{gray!18} 341,310
& \cellcolor{gray!18} --
& \cellcolor{gray!18} 5400 \\

& Gemini 3.1 Pro
& \textbf{0.070}
& 0.024
& 0.209
& 43
& 37
& 10.59
& --
& --
& 469 \\

& \cellcolor{gray!18} MiniMax M3
& \cellcolor{gray!18} 0.167
& \cellcolor{gray!18} 0.149
& \cellcolor{gray!18} 0.333
& \cellcolor{gray!18} 431
& \cellcolor{gray!18} 156
& \cellcolor{gray!18} 23.87
& \cellcolor{gray!18} 115,309
& \cellcolor{gray!18} --
& \cellcolor{gray!18} 6287 \\

\midrule

Full 3D Interaction
& \cellcolor{gray!18} Fable 5
& \cellcolor{gray!18} 0.043
& \cellcolor{gray!18} \textbf{0.005}
& \cellcolor{gray!18} 0.085
& \cellcolor{gray!18} 42
& \cellcolor{gray!18} 18
& \cellcolor{gray!18} 7.44
& \cellcolor{gray!18} 154,278
& \cellcolor{gray!18} 6.52
& \cellcolor{gray!18} 1064 \\

& Opus 5
& \textbf{0.015}
& 0.015
& \textbf{0.058}
& 80
& 34
& 15.18
& 380,011
& 7.92
& 1962 \\

& \cellcolor{gray!18} GPT-5.6 Sol
& \cellcolor{gray!18} 0.090
& \cellcolor{gray!18} 0.006
& \cellcolor{gray!18} 0.139
& \cellcolor{gray!18} 70
& \cellcolor{gray!18} 35
& \cellcolor{gray!18} 12.31
& \cellcolor{gray!18} 50,484
& \cellcolor{gray!18} 4.54$^{*}$
& \cellcolor{gray!18} 1344 \\

& Kimi K3
& 0.040
& 0.017
& 0.176
& 75
& 25
& 11.72
& 54,086
& --
& 2460 \\

& \cellcolor{gray!18} Qwen 3.8 Max
& \cellcolor{gray!18} 0.058
& \cellcolor{gray!18} 0.031
& \cellcolor{gray!18} 0.198
& \cellcolor{gray!18} 257
& \cellcolor{gray!18} 51
& \cellcolor{gray!18} 8.88
& \cellcolor{gray!18} 456,831
& \cellcolor{gray!18} --
& \cellcolor{gray!18} 8206 \\

& Gemini 3.1 Pro
& 0.036
& 0.006
& 0.083
& 38
& 22
& 6.45
& --
& --
& 448 \\

& \cellcolor{gray!18} MiniMax M3
& \cellcolor{gray!18} 0.113
& \cellcolor{gray!18} 0.058
& \cellcolor{gray!18} 0.250
& \cellcolor{gray!18} 172
& \cellcolor{gray!18} 67
& \cellcolor{gray!18} 10.59
& \cellcolor{gray!18} 76,278
& \cellcolor{gray!18} --
& \cellcolor{gray!18} 1905 \\

\bottomrule
\end{tabular}%
}
\label{tab:supp_main_results}
\end{table*}

\paragraph{Ablations.}
The pose/grid ablation uses the 10-instance subset with GPT-5.6 Sol under the
Multi-view protocol with three refinement iterations. \emph{Grid} replaces the
four target renders with native Blender viewport captures of the ground-truth
asset with the world grid enabled (major spacing 1 and minor spacing 0.25
scene units), taken from the same benchmark cameras at $512\times512$ in
Material Preview shading; \emph{Pose} keeps the original renders and appends
each view's camera metadata (azimuth, elevation, world-space camera location,
and a 50\,mm lens on a 36\,mm sensor) to the prompt. The observation-interface ablation
uses 10 instances with Gemini 3.1 Pro and MiniMax M3: \emph{Normalized Active
Visual} normalizes the viewport before every screenshot by switching to
\textsc{Object} mode, clearing the active object, disabling overlays, and
enforcing \textsc{Rendered} shading, while \emph{Fable 5 Inspection}
supplies the inspection screenshots collected by Fable 5 under Active Visual
(13.3 images per instance on average). The geometric-measurement ablation
uses 10 instances with Opus 5, adding to the Multi-view input global
measurements (bounding box, center, dimensions) and/or part-level measurements
(anonymous part centers and dimensions, component and repetition statistics)
extracted from the ground-truth mesh by a GPT-5.6 Sol-authored measurement
program.

\paragraph{VLM judge.}
The hexagonal evaluation (Sec.~\ref{sec:supp_hexagonal}) uses Gemini 3.1 Pro as the
judge, with a six-criterion rubric adapted from
GPTEval3D~\cite{wu2024gpteval3d} to absolute ground-truth fidelity; the exact
judge prompt is reproduced in Sec.~\ref{sec:supp_prompts}. For each instance,
a single request presents the $2\times2$ contact sheet of the four canonical
ground-truth renders followed by the contact sheets of all seven
reconstructions, anonymized and shuffled with a per-instance seed, and the
judge returns for every candidate an integer score from 1 to 5 on six
criteria: \emph{asset alignment}
(identity, silhouette, major part layout,
and basic proportions), \emph{3D plausibility} (a coherent 3D shape without
structural absurdities), \emph{geometry--texture alignment} (consistency
between geometry and materials, plausible shading, and cross-view coherence),
\emph{texture detail} (local texture, pattern, and color detail and
continuity), \emph{geometry detail} (shape fidelity, edge and curvature
detail, and secondary geometry), and \emph{fidelity} (one holistic score for
alignment with the ground truth, which we also report as the overall score).
Scores are averaged over the 100 instances of the Full 3D Interaction harness
for each of the seven agents; every reconstruction produced valid renders, so
no instance had to be scored as missing.

\section{Main Results}
\label{sec:supp_results}

\subsection{Topology and Interaction Efficiency}
\label{sec:supp_efficiency}

Table~\ref{tab:supp_main_results} reports the per-Betti-number topology
errors and the interaction-efficiency statistics defined in
Sec.~\ref{sec:supp_settings} for all agents and harnesses.

\paragraph{Topology improves with richer target access.}
Table~\ref{tab:supp_main_results} provides a per-Betti-number breakdown of the topology metric reported in the main paper.
The overall trend is consistent with the geometric results: richer target access generally enables agents to recover more faithful object structure.
In particular, Full 3D Interaction reduces the normalized error of all three Betti numbers relative to Single-view for every evaluated agent.
The improvement is especially pronounced for $\beta_0$ and $\beta_1$, suggesting that direct interaction with the target helps agents better recover both component structure and topological connectivity.
For example, Opus~5 reduces its $\beta_0$ error from 0.140 under Single-view to 0.015 under Full 3D Interaction, while GPT-5.6 Sol reduces its $\beta_1$ error from 0.097 to 0.006.
Active Visual exhibits a less uniform pattern: it substantially improves topology for several agents, but can also degrade individual Betti-number errors, reinforcing the observation from the main paper that additional visual access is useful only when an agent can effectively control and exploit its observations.

\paragraph{Interactive reconstruction exposes large efficiency differences across agents.}
The interaction statistics reveal substantially greater cross-agent variation
under Active Visual and Full 3D Interaction than under the fixed-view harnesses.
Single-view and Multi-view require relatively few reconstruction updates and no
target-side MCP interaction, whereas the interactive settings introduce repeated
cycles of target inspection, tool use, program revision, and comparison.

Agents use this interaction budget very differently. Under Active Visual,
MiniMax M3 performs the most extensive interaction, with 431 model API calls,
156 MCP calls, and 23.87 reconstruction versions on average, and also incurs
the highest latency. Qwen 3.8 Max similarly makes a large number of model calls,
while Opus 5 uses substantially fewer calls but generates more output tokens.
These differences show that interaction cost is multi-dimensional: more tool
calls, more reconstruction revisions, more generated tokens, and longer runtime
do not necessarily increase together. Overall, the interactive harnesses expose
not only differences in reconstruction quality, but also substantial variation
in how agents allocate computation and interaction to acquire and exploit
target-side evidence.

\begin{table*}[t]
    \centering
    \caption{
    \textbf{Comparison with 3DCodeBench on shared models.}
    We report the four target-access settings of 3DHarnessBench together
    with the four-view image-to-3D results from 3DCodeBench.
    The 3DCodeBench results provide a contextual fixed-view reference
    rather than a directly comparable baseline, since the reconstruction
    and evaluation protocols differ.
    $^{\dagger}$Chamfer values are particularly not directly comparable:
    3DCodeBench minimizes over four yaw rotations, whereas our evaluation
    minimizes over all 24 right-handed axis-aligned rotations.
    }
    \label{tab:3dcodebench_results}

    \setlength{\tabcolsep}{6pt}
    \renewcommand{\arraystretch}{1.10}

    \begin{tabular}{lllcccc}
        \toprule
        Model
        & Benchmark
        & Setting
        & SigLIP-2 $\uparrow$
        & DINOv3 $\uparrow$
        & Chamfer$^{\dagger}$ $\downarrow$
        & Uni3D $\uparrow$ \\
        \midrule

        \multirow{5}{*}{Fable 5}
        & 3DCodeBench
        & Image-to-3D (4 views)
        & 0.771
        & 0.538
        & 0.061
        & 0.558 \\
        \cmidrule(lr){2-7}

        & \multirow{4}{*}{3DHarnessBench}
        & Single-view (3 iterations)
        & 0.913
        & 0.784
        & 0.012
        & 0.733 \\

        &
        & Multi-view (3 iterations)
        & 0.927
        & 0.811
        & 0.011
        & 0.764 \\

        &
        & Active Visual
        & 0.932
        & 0.822
        & 0.006
        & 0.832 \\

        &
        & Full 3D Interaction
        & 0.949
        & 0.859
        & 0.003
        & 0.884 \\

        \midrule

        \multirow{5}{*}{Gemini 3.1 Pro}
        & 3DCodeBench
        & Image-to-3D (4 views)
        & 0.823
        & 0.540
        & 0.079
        & 0.518 \\
        \cmidrule(lr){2-7}

        & \multirow{4}{*}{3DHarnessBench}
        & Single-view (3 iterations)
        & 0.904
        & 0.756
        & 0.016
        & 0.728 \\

        &
        & Multi-view (3 iterations)
        & 0.908
        & 0.774
        & 0.015
        & 0.753 \\

        &
        & Active Visual
        & 0.865
        & 0.674
        & 0.015
        & 0.691 \\

        &
        & Full 3D Interaction
        & 0.901
        & 0.759
        & 0.008
        & 0.783 \\

        \bottomrule
    \end{tabular}
\end{table*}

\paragraph{Richer geometric access does not necessarily require more interaction.}
Interestingly, moving from Active Visual to Full 3D Interaction lowers the
number of model API and MCP API calls for most agents (GPT-5.6 Sol's API calls
are the exception, rising from 55 to 70), despite exposing a strictly richer
target-side interface.

Part of this reduction is explained by the interaction protocol itself.
Active Visual maintains the target and reconstruction in separate Blender
processes connected through different MCP interfaces. Comparing the two
therefore often requires separate viewport-control, alignment, and screenshot
operations on the target and reconstruction sides. In Full 3D Interaction,
both the target and reconstruction are in the same Blender process and
are accessed through a single MCP interface, reducing this coordination
overhead.

Beyond this protocol-level difference, explicit geometric queries may also
provide more informative evidence per interaction than repeated visual
inspection, reducing the need for additional exploration and refinement.
The lower interaction counts therefore reflect more efficient evidence acquisition.

\begin{figure}[t]
    \centering
    \includegraphics[width=\linewidth]
    {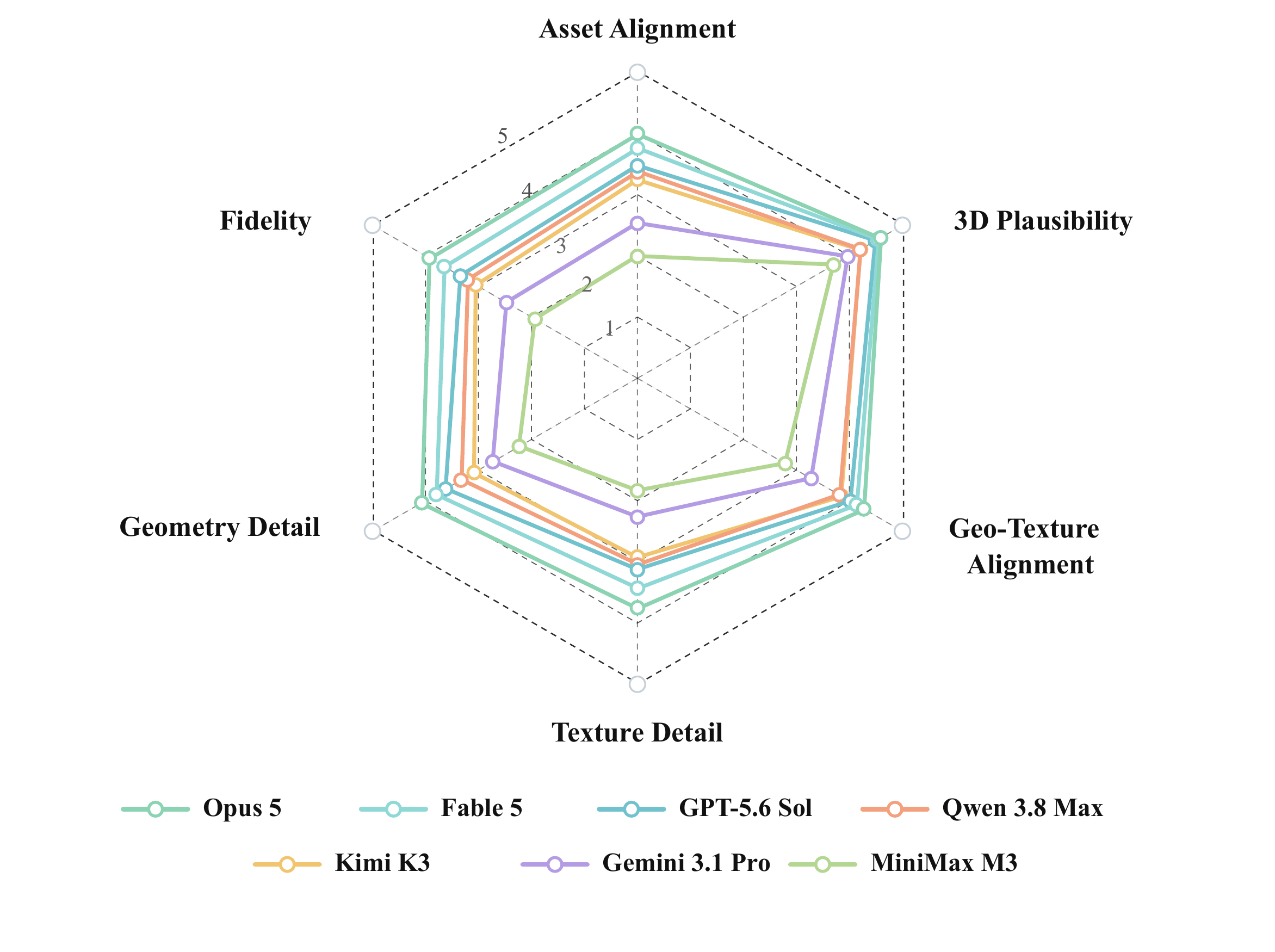}
    \caption{\textbf{VLM-based reconstruction evaluation across seven agents.}
Gemini 3.1 Pro, used as the judge, compares each Full 3D Interaction
reconstruction (100 instances per agent) against the corresponding
ground-truth renders and assigns an integer score from 1 to 5 on each of the
six criteria defined in Sec.~\ref{sec:supp_settings}.
}
    \label{fig:hexalgonal}
\end{figure}

\subsection{Comparison with 3DCodeBench}
\label{sec:supp_3dcodebench}

\begin{table*}[t]
\centering
\caption{\textbf{Pose and grid cues in Multi-view reconstruction} (10
instances, GPT-5.6 Sol). Camera poses give the most consistent gains; grid
cues help less reliably, and combining both is not additive. }
\label{tab:supp_pose_grid}
% \resizebox{\columnwidth}{!}{%
\begin{tabular}{lccccccc}
\toprule
Setting
& SIGLIP-2 $\uparrow$
& DINOv2 $\uparrow$
& DINOv3 $\uparrow$
& Chamfer $\downarrow$
& UNI3D $\uparrow$
& UNI3D I--3D $\uparrow$
& Betti Norm. L1 $\downarrow$ \\
\midrule
Baseline
& 0.9254
& 0.7238
& 0.7765
& 0.0199
& 0.7038
& 0.3057
& 0.7567 \\

+ Grid
& 0.9321
& 0.7445
& 0.7974
& 0.0172
& 0.8351
& \textbf{0.3754}
& \textbf{0.7487} \\

+ Pose
& \textbf{0.9419}
& \textbf{0.7672}
& \textbf{0.8107}
& \textbf{0.0077}
& \textbf{0.8387}
& 0.3621
& 0.8071 \\

+ Grid + Pose
& 0.9391
& 0.7467
& 0.7868
& 0.0110
& 0.8098
& 0.3561
& 0.9104 \\
\bottomrule
\end{tabular}%
% }
\end{table*}

\begin{table*}[t]
\centering
\caption{\textbf{Explicit geometric measurements added to Multi-view input}
(10 instances, Opus 5). All measurement settings improve geometry over
image-only input, and global and part-level measurements are complementary,
with their combination achieving the lowest Chamfer. }
\label{tab:supp_measurement_ablation}
% \resizebox{\columnwidth}{!}{%
\scriptsize
\begin{tabular}{lccccccc}
\toprule
Setting
& SIGLIP-2 $\uparrow$
& DINOv2 $\uparrow$
& DINOv3 $\uparrow$
& Chamfer $\downarrow$
& UNI3D $\uparrow$
& UNI3D I--3D $\uparrow$
& Betti Norm. L1 $\downarrow$ \\
\midrule
Imgs
& 0.9173
& 0.6847
& 0.7404
& 0.0150
& 0.8133
& 0.3625
& 0.9050 \\

Imgs + Global
& \textbf{0.9309}
& \textbf{0.7171}
& 0.7410
& 0.0103
& 0.8088
& 0.3537
& 0.7093 \\

Imgs + Parts
& 0.9256
& 0.7109
& \textbf{0.7790}
& 0.0099
& \textbf{0.8557}
& 0.3498
& \textbf{0.4186} \\

Imgs + Global + Parts
& 0.9307
& 0.6914
& 0.7561
& \textbf{0.0084}
& 0.8296
& \textbf{0.3645}
& 0.5783 \\
\bottomrule
\end{tabular}
% }
\end{table*}

3DCodeBench~\cite{gao2026_3dcodebench} is the closest prior benchmark to our
setting: both evaluate executable 3D reconstruction through Blender Python
programs, and our targets are drawn from it. As Table~\ref{tab:evaluation_protocol}
shows, our \textbf{Multi-view} setting is most closely related to its
four-view image-to-3D protocol---both provide four fixed target views and use
program execution and rendering as the reconstruction interface---but our
fixed-view settings additionally introduce three render-and-refine rounds,
allowing agents to visually inspect and revise their reconstructions after
successful execution. Active Visual and Full 3D Interaction then extend this
fixed-view protocol toward agentic interaction through Blender MCP: rather than
replacing the fixed-image setting of 3DCodeBench, 3DHarnessBench uses it as a
starting point and progressively evaluates an agent's ability to actively
acquire and exploit richer target-side evidence.

\begin{table*}[t]
\centering
\caption{\textbf{Observation-interface ablation under Active Visual} (10
instances). Normalized viewport screenshots help MiniMax M3 across most metrics,
while their effect on Gemini 3.1 Pro is mixed; Fable 5's inspection views help
Gemini 3.1 Pro but not MiniMax M3, showing that acquiring and exploiting
observations are both agent-dependent.}
\label{tab:supp_observation_ablation}
\scriptsize
% \setlength{\tabcolsep}{2.0pt}
% \renewcommand{\arraystretch}{0.95}

% \resizebox{\columnwidth}{!}{%
\begin{tabular}{llccccccc}
\toprule
\textbf{Model}
& \textbf{Setting}
& \textbf{SIGLIP-2} $\uparrow$
& \textbf{DINOv2} $\uparrow$
& \textbf{DINOv3} $\uparrow$
& \textbf{Chamfer} $\downarrow$
& \textbf{UNI3D} $\uparrow$
& \textbf{UNI3D I--3D} $\uparrow$
& \textbf{Betti Norm. L1} $\downarrow$ \\
\midrule

\multirow{4}{*}{Gemini 3.1 Pro}
& Multi-view
& 0.9189
& 0.6224
& \textbf{0.7212}
& \textbf{0.0165}
& \textbf{0.7609}
& 0.3471
& 0.6837 \\

& Active Visual
& 0.8452
& 0.4995
& 0.6031
& 0.0234
& 0.6776
& 0.3172
& \textbf{0.6116} \\

& Normalized Active Visual
& 0.8710
& 0.5499
& 0.6229
& 0.0186
& 0.7321
& 0.3232
& 0.9065 \\

& Fable 5 Inspection
& \textbf{0.9232}
& \textbf{0.6307}
& 0.7164
& 0.0191
& 0.7605
& \textbf{0.3549}
& 0.6793 \\

\midrule

\multirow{4}{*}{MiniMax M3}
& Multi-view
& 0.8230
& 0.3777
& \textbf{0.5053}
& 0.0262
& \textbf{0.6200}
& 0.2799
& 0.9696 \\

& Active Visual
& 0.7583
& 0.2926
& 0.3917
& 0.0798
& 0.3722
& 0.2137
& 0.9863 \\

& Normalized Active Visual
& \textbf{0.8315}
& \textbf{0.4016}
& 0.5015
& \textbf{0.0240}
& 0.6148
& \textbf{0.3074}
& \textbf{0.8326} \\

& Fable 5 Inspection
& 0.8279
& 0.3792
& 0.4733
& 0.0414
& 0.4899
& 0.2601
& 0.9541 \\

\bottomrule
\end{tabular}%
% }
\end{table*}
\begin{table}[t]
\centering
\caption{
\textbf{Active Visual inspection behavior across agents.}
We report the average number of target-inspection calls,
reconstruction-comparison calls, and their total under Active Visual,
computed from the screenshot-producing Blender MCP calls on the target
side (\texttt{get\_viewport\_screenshot} of the restricted server) and on the
reconstruction side (\texttt{get\_screenshot\_*} of the official server).
Gemini 3.1 Pro performs substantially fewer inspection and comparison calls
than the other agents, consistent with its tendency to under-explore the
target and terminate the interaction early.
}
\label{tab:active_visual_tool_calling}
\resizebox{\columnwidth}{!}{%
\begin{tabular}{lccc}
\toprule
\textbf{Agent}
& \textbf{GT Inspection}
& \textbf{Comparison}
& \textbf{Total} \\
\midrule

GPT-5.6 Sol
& 13.15 & 12.58 & 25.73 \\

Fable 5
& 10.28 & 2.59 & 12.87 \\

Opus 5
& 17.76 & 8.39 & 26.15 \\

Gemini 3.1 Pro
& 5.42 & 2.58 & 8.00 \\

Kimi K3
& 10.17 & 4.90 & 15.07 \\

Qwen 3.8 Max
& 6.10 & 5.18 & 11.28 \\

MiniMax M3
& 31.71 & 43.11 & 74.82 \\

\bottomrule
\end{tabular}%
}
\end{table}

\paragraph{Results on shared models.}
Table~\ref{tab:3dcodebench_results} places our four settings next to the
four-view image-to-3D results reported by 3DCodeBench for the two models
common to both benchmarks, Fable~5 and Gemini~3.1~Pro. The absolute numbers
are not directly comparable: we evaluate a 100-object subset, our fixed-view
settings add three render-and-refine rounds, and the Chamfer protocols differ
(four yaw rotations in 3DCodeBench versus all 24 axis-aligned rotations here),
which is why our fixed-view scores are uniformly higher. The comparison is
nevertheless informative in two ways. First, the geometric ordering of the
shared models is preserved: Fable~5 attains lower Chamfer distance and higher
Uni3D similarity than Gemini~3.1~Pro in both benchmarks, whereas the
appearance metrics favor Gemini~3.1~Pro in 3DCodeBench but Fable~5 under our
Multi-view setting. Second, the two models diverge sharply once target access
becomes agent-directed: from Multi-view to Active Visual, Fable~5 improves
from $0.764$ to $0.832$ Uni3D while Gemini~3.1~Pro drops from $0.753$ to
$0.691$, a distinction that no fixed-image protocol can expose.

\subsection{VLM-based Hexagonal Evaluation}
\label{sec:supp_hexagonal}

Figure~\ref{fig:hexalgonal} provides a complementary perceptual evaluation of
the Full 3D Interaction reconstructions using Gemini 3.1 Pro as a judge,
following the protocol and the six criteria defined in
Sec.~\ref{sec:supp_settings}. The resulting ranking is broadly consistent
across the six criteria: Opus 5 achieves the strongest overall performance,
followed by Fable 5 and GPT-5.6 Sol, while Gemini 3.1 Pro and MiniMax M3
remain weaker. The judge shares its base model with one of the evaluated
agents; since Gemini 3.1 Pro's own reconstructions nevertheless receive the
second-lowest scores, judge self-preference does not appear to drive the
ranking.

A notable pattern is that the agents are more closely clustered on
3D plausibility than on fidelity, geometry detail, and texture detail.
Even weaker agents can often produce a coherent and plausible 3D object,
but they are substantially less successful at matching the identity,
proportions, and fine-grained structure of the ground truth.
This suggests that the main remaining challenge is not merely generating a
valid 3D shape, but reconstructing the specific target faithfully and with
sufficient geometric and appearance detail.

\section{Ablations and Analyses}
\label{sec:supp_ablation_results}

The main paper reports a compact subset of metrics for readability. This
section provides the complete results, additionally including DINOv2, DINOv3,
and Uni3D image--3D similarity (Uni3D I--3D), and extends the analyses with
tool-calling statistics and qualitative trajectories. The additional metrics
broadly support the conclusions in the main paper, while also revealing that
changes in geometric fidelity and visual similarity are not always aligned
across different feature spaces.

\subsection{Geometric Grounding in Multi-view}
\label{sec:supp_pose_grid}

\paragraph{Pose and grid cues.}
Table~\ref{tab:supp_pose_grid} reports the complete results for the pose and
grid ablation. Camera pose remains the most reliable source of geometric
grounding, improving both geometry and several appearance metrics.
Grid cues provide less consistent gains, although they can improve Uni3D
I--3D similarity. Combining grid and pose is not additive and can degrade
some appearance and topology metrics.
Overall, the additional metrics reinforce that pose information is the more
useful and robust geometric cue.

\paragraph{Explicit geometric measurements.}
\label{sec:supp_measurement_ablation}
Table~\ref{tab:supp_measurement_ablation} reports the complete geometric
measurement ablation. Explicit measurements improve most metrics over
image-only input, with particularly consistent gains in Chamfer distance and
topology. Part-level measurements achieve the best Uni3D and topology, while
combining global and part-level cues yields the lowest Chamfer and highest
Uni3D I--3D. The remaining metric variation suggests that global and part-level
measurements provide complementary information, with their clearest benefits
in geometric fidelity and topology.

\begin{figure*}[t]
    \centering
    \includegraphics[width=0.8\linewidth]{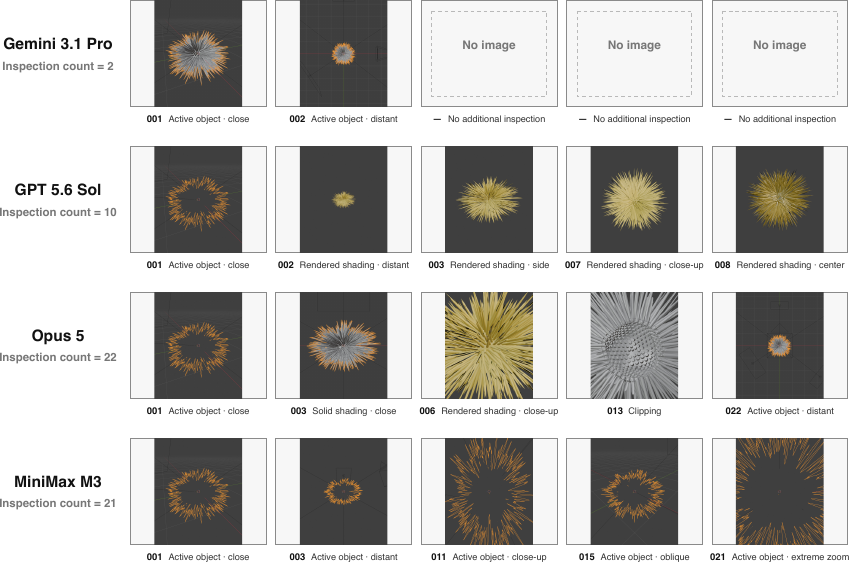}
    \caption{
        \textbf{Urchin inspection trajectories under Active Visual.}
        Selected observations reveal large agent-dependent differences in inspection
        frequency and diversity. Gemini 3.1 Pro terminates after very limited
        inspection, GPT-5.6 Sol and Opus 5 explore complementary viewpoints and
        rendering modes, while MiniMax M3 performs many but relatively repetitive
        active-object inspections. Inspection counts denote the total number of target
        inspections for this instance.
    }
    \label{fig:active_visual_inspect}
\end{figure*}

\subsection{Active Visual}
\label{sec:supp_active_visual}

Under the Active Visual setting, Gemini 3.1 Pro and MiniMax M3 exhibit
performance degradation relative to Multi-view, whereas the other agents
generally benefit from richer target access. This subsection investigates the
sources of this degradation through a controlled ablation, tool-calling
statistics, and qualitative visualizations, focusing on how observation
quality, inspection behavior, and the ability to exploit acquired visual
evidence affect Active Visual performance.

\paragraph{Observation-interface ablation.}
\label{sec:supp_observation_ablation}
Table~\ref{tab:supp_observation_ablation} reports the complete Active Visual
observation ablation. The additional metrics confirm that standard Active
Visual substantially degrades reconstruction quality for both weaker agents.
Viewport normalization recovers much of this degradation for MiniMax M3,
improving all reported metrics relative to standard Active Visual and even
surpassing its Multi-view baseline on several metrics. For Gemini 3.1 Pro,
normalization improves appearance and geometry metrics but substantially
worsens topology, indicating a more mixed effect.

Providing Fable 5 inspection views strongly benefits Gemini 3.1 Pro, recovering
or nearly matching its Multi-view performance across most metrics. MiniMax M3
also improves over standard Active Visual with the same inspection views, but
remains well below its Multi-view baseline on most metrics and benefits more
from viewport normalization. These results suggest different failure modes:
Gemini is particularly limited by the observations it acquires, whereas
MiniMax M3 is affected by both observation quality and its ability to exploit
informative observations.

\paragraph{Inspection tool-calling analysis.}
Table~\ref{tab:active_visual_tool_calling} breaks the agents' Active Visual
MCP usage into target-inspection calls (screenshots of the target acquired
through the constrained target-side interface) and reconstruction-comparison
calls (screenshots of the current reconstruction taken for comparison); the
remaining MCP calls counted in Table~\ref{tab:supp_main_results} are code
execution and scene queries, which is why the totals here are lower than the
MCP totals there. Agents adopt markedly different interaction budgets.
Gemini 3.1 Pro is the most conservative, issuing only 5.42 target-inspection
calls and 2.58 reconstruction-comparison calls on average, for a total of 8.00.
This is substantially lower than the other agents and is consistent with its
tendency to terminate the interaction after acquiring only limited visual
evidence. At the other extreme, MiniMax M3 performs 31.71 inspections and
43.11 comparisons on average, reaching 74.82 total calls, which indicates a
much more exhaustive but also substantially less economical interaction
strategy. GPT-5.6 Sol and Opus 5 occupy an intermediate regime, with roughly
26 total calls, while Fable 5 and Qwen 3.8 Max use comparatively compact
interaction budgets.

These results suggest that insufficient interaction can leave important
geometric or appearance ambiguities unresolved, thereby degrading reconstruction
performance. This interpretation is also consistent with the observation
ablation above: when Gemini is provided with inspection views collected by
Fable 5, its reconstruction quality improves substantially. Thus, a successful
Active Visual agent must not only interpret observations correctly, but also
learn when and how to acquire informative, high-quality observations
efficiently.

\paragraph{Inspection trajectories.}
Figure~\ref{fig:active_visual_inspect} further reveals substantial differences
in inspection strategy across agents. Gemini 3.1 Pro performs only two
inspections before terminating, leaving limited opportunity to resolve remaining
ambiguities. In contrast, GPT-5.6 Sol and Opus 5 explore more diverse viewpoints
and rendering modes. Notably, Opus 5 also adjusts the camera clipping plane to
slice through the target, using the viewport to expose otherwise occluded
geometry rather than only inspecting the exterior surface.

MiniMax M3 performs many inspections, but these are largely repeated
active-object views with varying zoom and orientation, providing less diverse
evidence. Together with the tool-call statistics in
Table~\ref{tab:active_visual_tool_calling}, these examples suggest that effective
active inspection depends not only on the amount of inspection, but also on
the diversity, quality, and informativeness of the acquired observations.

\begin{figure*}[t]
    \centering
    \includegraphics[width=0.9\linewidth]{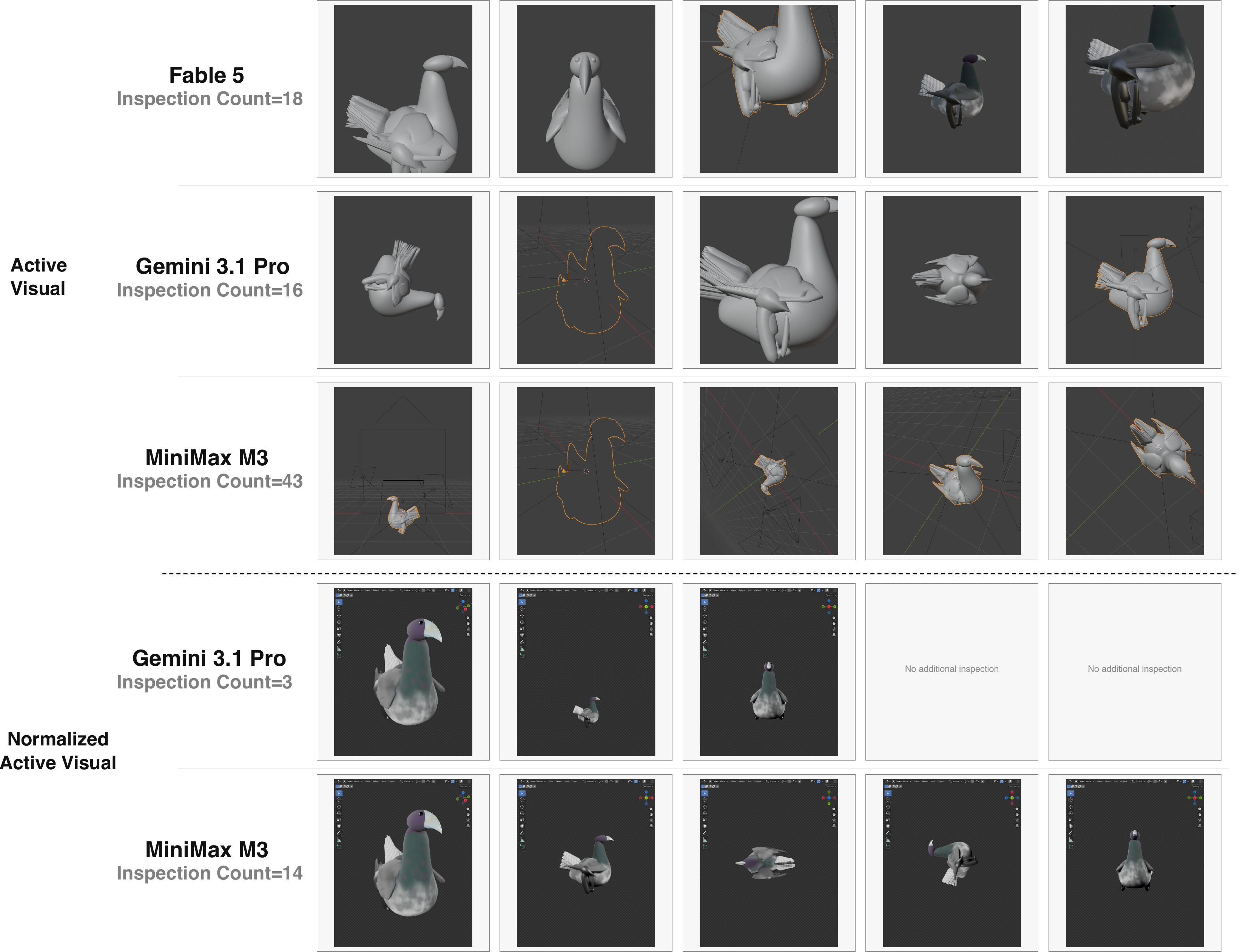}
    \caption{\textbf{Inspection trajectories for the bird example under Active Visual ablations.} Rows show selected target observations acquired by Fable 5, Gemini 3.1 Pro, and MiniMax M3 under standard Active Visual and Normalized Active Visual; inspection counts report the total number of target inspections for this
instance. The Fable 5 Active Visual trajectory is also used as the source for the \emph{Fable 5 Inspection} ablation setting. Viewport normalization
produces more consistently framed observations, while the trajectories reveal
large agent-dependent differences in both inspection number and observation
quality.}
    \label{fig:analysis_ablation_observation}
\end{figure*}

\paragraph{Visualization of the observation-interface ablation.}
Figure~\ref{fig:analysis_ablation_observation} visualizes inspection trajectories
for the bird example under the Active Visual observation-interface ablation.
The Fable 5 row shows the inspection trajectory collected by Fable 5 under
the standard Active Visual setting; these observations are subsequently reused
as the fixed inspection images in the \emph{Fable 5 Inspection} ablation.
\emph{Normalized Active Visual}, in contrast, preserves each agent's own
inspection decisions while normalizing the viewport state before every
screenshot to provide more consistent observations.

The examples illustrate that Active Visual performance is jointly determined
by inspection frequency and observation quality. Under standard Active Visual,
MiniMax M3 performs many inspections, but several views are poorly framed,
distant, or affected by suboptimal viewport states. Normalization produces
substantially more consistent, target-centered observations despite fewer
inspection calls. Gemini 3.1 Pro, on the other hand, performs only a few
inspections after normalization and terminates early, consistent with the
limited-exploration behavior discussed previously. Together, these examples
show that effective Active Visual reconstruction requires both adequate
exploration and informative observations.

\subsection{Initialization versus Interactive Refinement}
\label{sec:supp_trajectories}

\begin{figure*}[t]
    \centering
    \includegraphics[width=\linewidth]{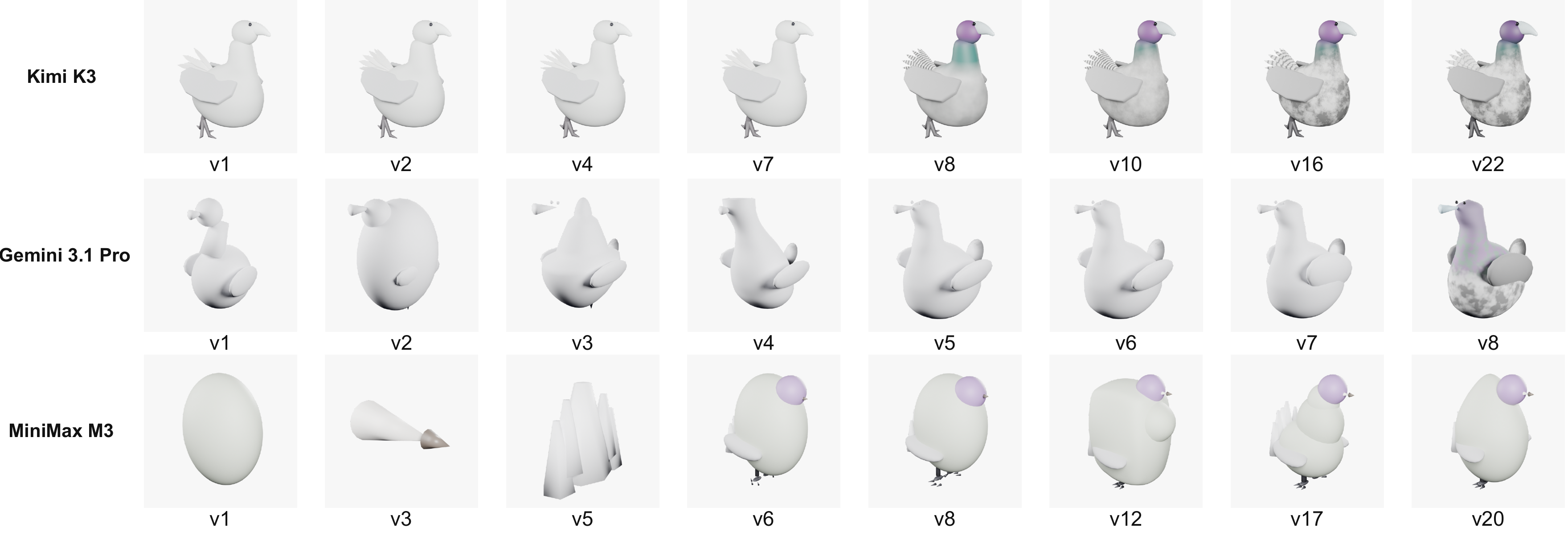}
    \caption{\textbf{Reconstruction trajectories under Full 3D
    Interaction.}
    Selected reconstructed versions from three agents on the
    same example illustrate substantial differences in trajectory length and
    in how the reconstruction changes over self-correction.}
    \label{fig:version}
\end{figure*}

\begin{figure*}[h]
    \centering
    \includegraphics[width=0.8\linewidth]
    {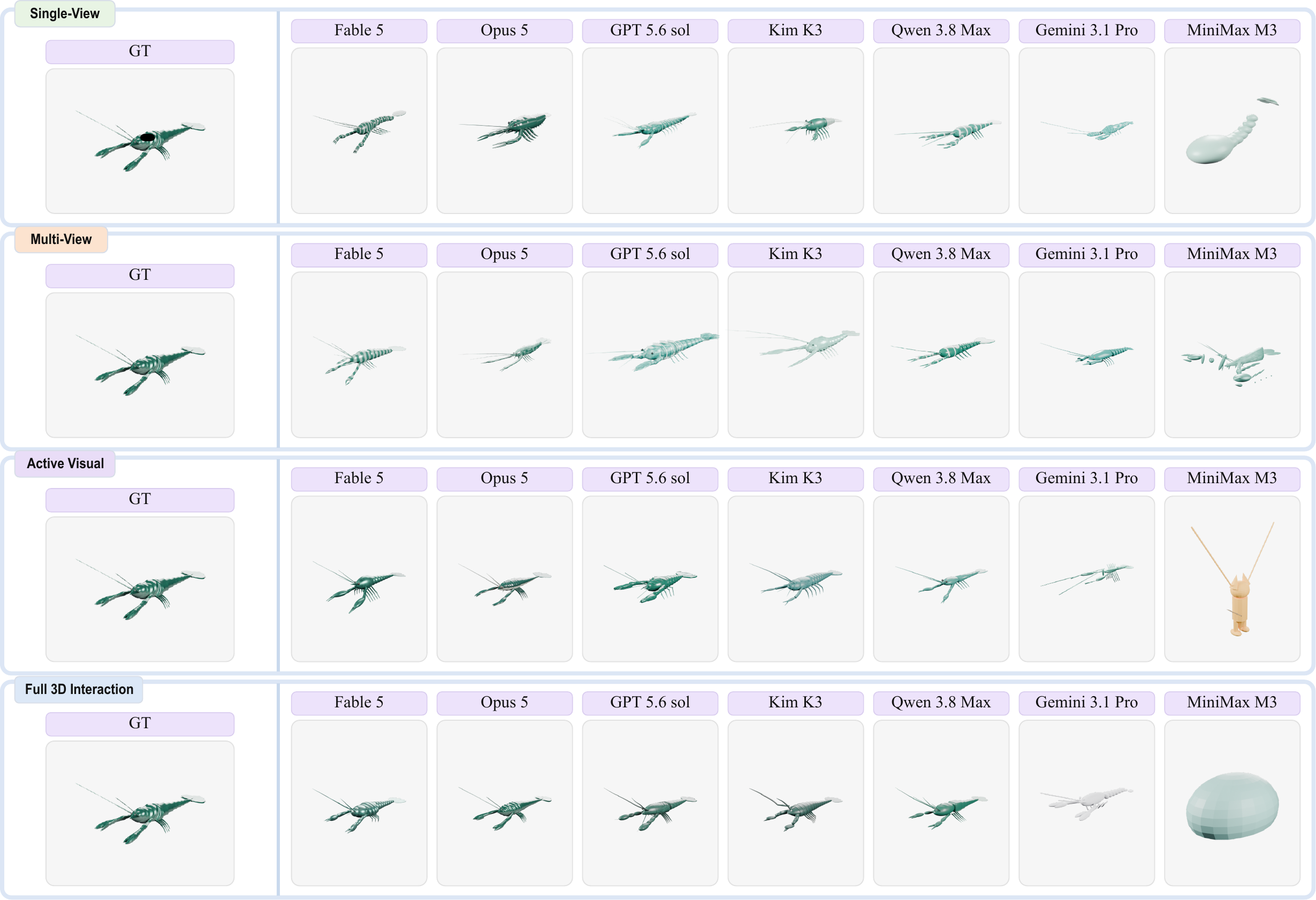}
    \caption{
\textbf{Crustacean reconstruction across four target-access settings.}
Reconstructions from seven frontier VLM agents under Single-view, Multi-view,
Active Visual, and Full 3D Interaction. Richer target access helps stronger
agents recover the elongated body and articulated appendages more faithfully,
while substantial failures remain for weaker agents.
}
    \label{fig:vis_crustacean}
\end{figure*}

Figure~\ref{fig:version} visualizes intermediate
reconstruction versions for three agents on the same target.
The trajectories reveal qualitatively different refinement behaviors.
Kimi K3 establishes the overall object structure early and then progressively
adds appearance and local details over a relatively long trajectory.
Gemini 3.1 Pro follows a much shorter trajectory, making several geometric
revisions before terminating after only a few versions.
MiniMax M3 exhibits a different pattern: its early reconstructions are far
from the target structure, but later iterations substantially reorganize the
geometry and eventually recover a recognizable object.

These examples illustrate that interactive refinement is not simply local
polishing of an initial reconstruction. Agents differ in when they obtain a
usable 3D hypothesis, how extensively they revise it, and whether later
iterations refine an existing structure or correct earlier structural errors.
This further supports the finding that initialization quality and
self-correction are distinct agent capabilities.

\begin{figure*}[t]
    \centering
    \includegraphics[width=0.75\linewidth]
    {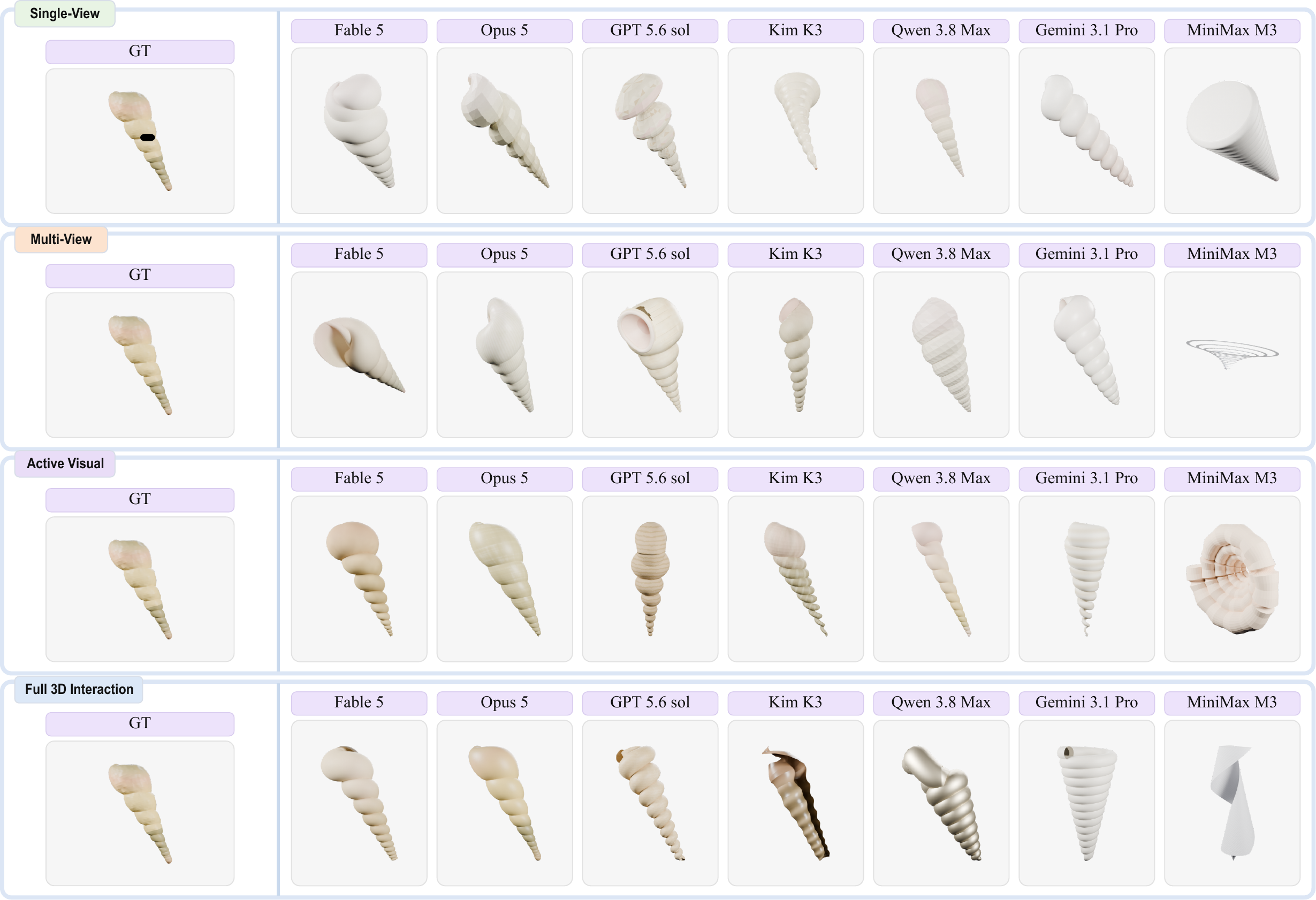}
    \caption{
\textbf{Auger reconstruction across four target-access settings.}
Reconstructions from seven frontier VLM agents under Single-view, Multi-view,
Active Visual, and Full 3D Interaction. Interactive access generally improves
recovery of the tapered global shape and repeated helical structure, while
cross-agent differences remain in structural fidelity.
}
    \label{fig:vis_auger}
\end{figure*}

\begin{figure*}[t]
    \centering
    \includegraphics[width=0.75\linewidth]
    {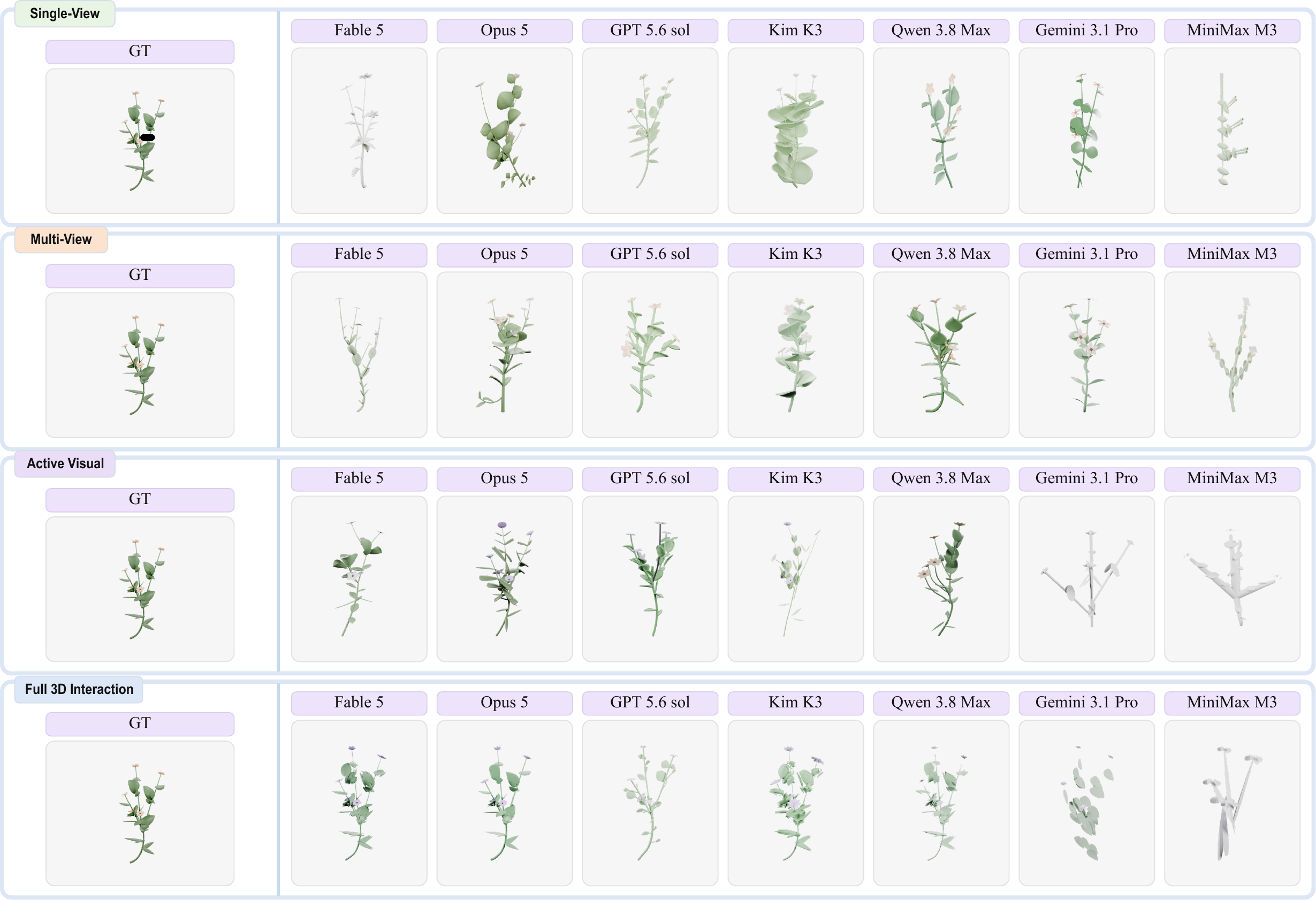}
    \caption{
\textbf{FlowerPlant reconstruction across four target-access settings.}
Reconstructions from seven frontier VLM agents under Single-view, Multi-view,
Active Visual, and Full 3D Interaction. Richer target access helps stronger
agents recover the multi-part arrangement of stems, leaves, and blossoms,
whereas weaker agents often simplify or omit fine structures.
}
    \label{fig:vis_flowerplant}
\end{figure*}

\begin{figure*}[t]
    \centering
    \includegraphics[width=\linewidth]
    {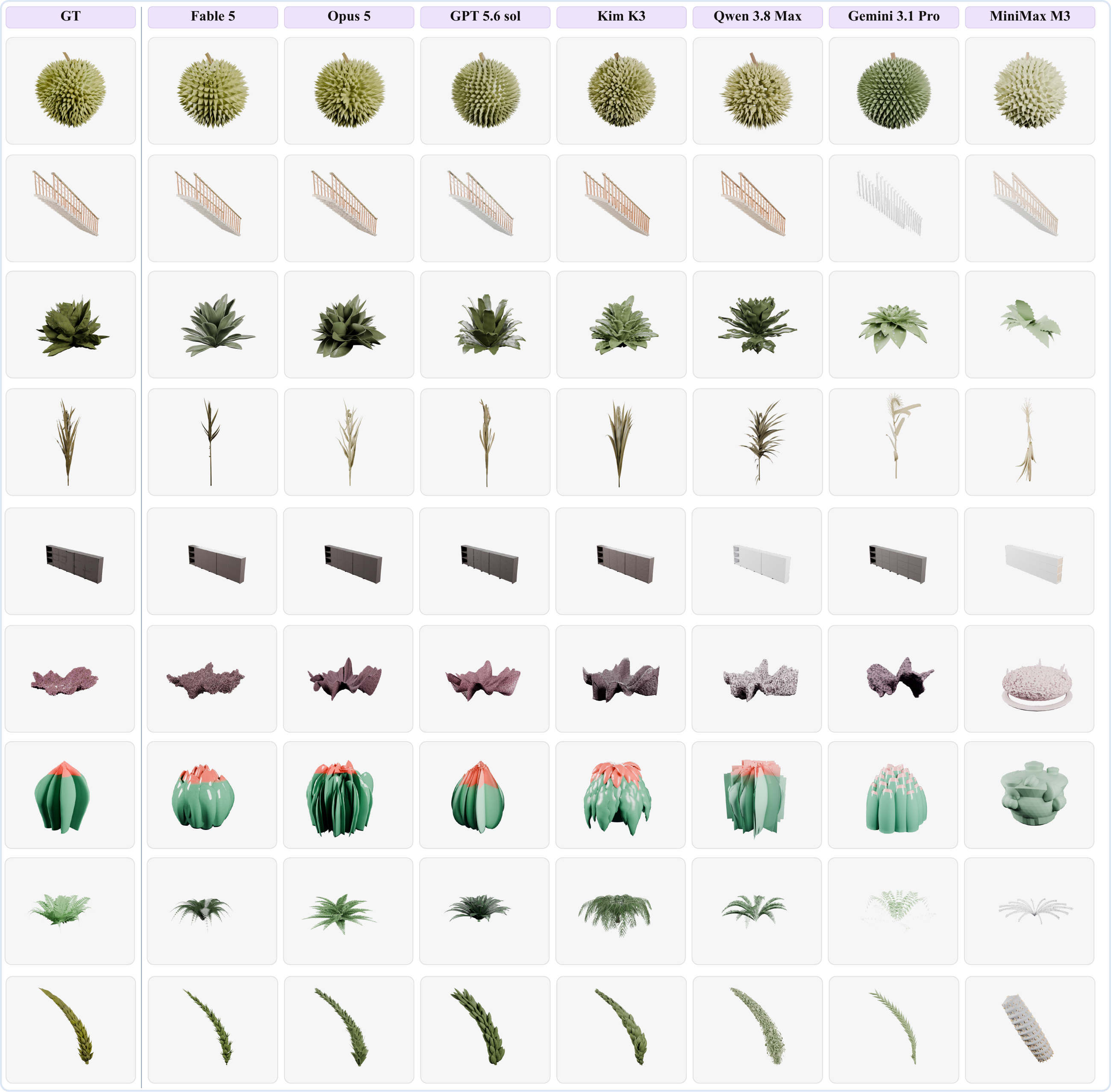}
    \caption{\textbf{Qualitative comparison under Full 3D Interaction.}
    Seven frontier VLM agents reconstruct the same set of target objects with
    access to the Full 3D Interaction interface. Substantial differences remain
    across agents, including visible differences on repeated, thin, and
    multi-part structures.}
    \label{fig:vis_full_mcp_9}
\end{figure*}

\section{Additional Qualitative Results}
\label{sec:supp_visualization}

\paragraph{Reconstructions across the four harnesses.}
Figures~\ref{fig:vis_crustacean}--\ref{fig:vis_flowerplant} provide additional
comparisons across the four harness settings.
The examples reinforce that richer target access is particularly useful for
objects with ambiguous, thin, repeated, or highly structured geometry.
For stronger agents, Active Visual and Full 3D Interaction generally recover
more complete part layouts and more faithful proportions than the fixed-view
settings. This is visible in the crustacean, where interactive access helps
recover the elongated body and articulated appendages; in the auger, where
it improves the tapered profile and repeated helical structure; and in the
flower plant, where it helps recover the arrangement of stems, leaves, and
blossoms.

At the same time, these examples illustrate why Active Visual is the strongest
discriminator between agents. Additional viewpoints can substantially improve
reconstruction for some agents, while others still produce simplified or
structurally incorrect geometry despite having access to agent-directed observations.
Full 3D Interaction is more robust overall, but large cross-agent differences
remain, especially for fine, repeated, and multi-part structures.

\paragraph{Cross-agent differences under Full 3D Interaction.}
Figure~\ref{fig:vis_full_mcp_9} further compares all seven agents under
Full 3D Interaction across a diverse set of targets.
Even with access to explicit geometric information, reconstruction quality
remains strongly agent-dependent. Stronger agents generally preserve the
global silhouette, repeated structures, and major part relationships more
faithfully, whereas weaker agents often simplify thin structures, miss
repeated components, or recover only the coarse object category.
The remaining errors are especially visible on objects with many small
repeated elements or delicate geometry, suggesting that full target access
reduces geometric ambiguity but does not eliminate the need for effective
structural reasoning and programmatic modeling.

% \clearpage
% \raggedbottom

\newpage

\section{Prompts and Agent Skill}
\label{sec:supp_prompts}

\paragraph{Single-view and Multi-view.}
The fixed-view harnesses adopt the initial reconstruction prompt from
3DCodeBench~\cite{gao2026_3dcodebench} and, for iterative refinement, the
editing prompt from BlenderGym~\cite{gu2025blendergym}. The exact prompt used for
the initial reconstruction is:
\clearpage
\begin{Verbatim}[fontsize=\scriptsize,breaklines=true,frame=single,framesep=2mm]

GT reference image(s) of one target object are attached. Inspect all of them before writing the reconstruction.

You are a procedural 3D modeling expert. Given one or more reference images of a 3D object, you produce a single self-contained Python script that reconstructs both the depicted geometry and its visible surface appearance, including materials, colors, patterns, roughness, metallic response, transparency, and other texture-related properties in Blender 5.0.

# Input format

You will receive one or more reference images of a single target object as part of the user message. The images may be:
- A single view (front, side, 3/4, etc.) -- infer unseen sides by symmetry and category priors.
- Multiple views of the SAME object (e.g. front + side + back, or turntable frames). Treat them as multi-view evidence of one object, not as separate objects. Cross-reference views to resolve depth, proportions, and occluded structure.
- A mix of full-object shots and close-up detail crops. Use the close-ups to refine local geometry such as handles, vents, and ornament on the same object.

If the images appear to depict different objects, model the most prominent or first-shown object and silently ignore the rest.

Read the images carefully before writing code. Identify the object's category, overall proportions such as length:width:height, part decomposition, symmetries such as bilateral, radial, or none, repeating structures such as slats, ribs, teeth, scales, or leaves, and any distinctive ornament. Your script should reproduce these as real geometry.

Also analyze the visible surface appearance. Identify material regions, dominant colors, color boundaries, repeating patterns, labels or markings represented visually, roughness, metallic response, transparency, translucency, emission, and significant bump or displacement. Use this evidence to reproduce the object's appearance with Blender materials and procedural or generated textures.

# Output format

Your entire response will be saved verbatim into a `.py` file and executed by Blender 5.0. Any non-Python content anywhere in the response will break that file. Treat this as a strict machine-to-machine contract, not as a chat reply.

Your response must consist of nothing but Python source code.

- Do not emit Markdown code fences.
- Do not prepend or append prose.
- Do not include HTML or XML tags, bullet points, headings, or formatting outside Python source.
- Record image observations only as Python comments inside the script.
- The first character of the response must be the first character of valid Python source.
- The final character of the response must be part of the Python script.

Before answering, ensure that this command succeeds without modification:

python -c "import ast; ast.parse(open('out.py').read())"

# Target environment

- Blender version: 5.0.
- Use the Blender 5.0 Python API through `bpy`, `bmesh`, and `mathutils`.
- The script will be executed with `blender --background --python <file>` or pasted into the Blender Text Editor.
- Allowed imports:
  - Blender-bundled: `bpy`, `bmesh`, `mathutils`
  - Python standard library: `math`, `random`, `itertools`, `collections`, `functools`, `dataclasses`, `enum`, `typing`
  - Numerical: `numpy`, `scipy`
- Do not import libraries requiring network access, GUI interaction, or external file reads.
- Do not import `os`, modify `sys.path`, or use `requests`, `PIL`, `cv2`, or similar libraries.
- The reference images are unavailable to the script at runtime. Do not attempt to load, open, or read them from disk. Encode everything inferred from the images directly into the generated scene.

# Code requirements

- Produce one final 3D object or coherent assembly corresponding to the object shown in the reference images.
- Generate only the depicted object. Do not generate a ground plane, backdrop, skybox, environmental props, decorative context, floor, stand, hand, or other surrounding scenery.
- Place the object at the origin.
- Match the reference as faithfully as geometry allows, including overall silhouette, part proportions, repeating-element counts, placement, and characteristic curvature.
- When the images provide clear quantitative cues such as eight spokes or three drawers, reproduce those counts exactly.
- Push geometric detail as far as the images warrant. Model ribs, slats, vents, handles, fins, scales, leaves, rivets, pleating, segmentation, and ornament as real geometry when they have visible depth.
- Use subdivision surface, bevel, array, mirror, screw, solidify, displacement, and other modifiers where appropriate.
- Use `bmesh` for local topology and fine geometric details where appropriate.
- Keep the final scene within a few hundred thousand vertices so execution remains within a 240-second budget.
- Prefer procedural construction with parametric loops, `bmesh` operators, modifiers, arrays, mirrors, and radial duplication.
- Avoid long hard-coded vertex lists.
- Exploit bilateral or radial symmetry when supported by the reference.
- At the beginning of the script, clear the default scene, including the default cube, camera, and light.

- Reconstruct both geometry and visible surface appearance. Match the reference object's material separation, dominant colors, patterns, roughness, metallic response, transparency, translucency, emission, and significant surface relief.
- Create materials through Blender's node system. Use procedural textures, generated image textures, vertex colors, UV coordinates, or geometry-based masks where appropriate.
- The reference images are unavailable at runtime. Encode all inferred colors, patterns, masks, and texture data directly in the script.
- Texture details that change the silhouette or create visible depth must still be modeled as geometry or displacement rather than represented only by color.

- Do not create a ground plane, lighting rig, camera, backdrop, or environment.
- Do not save the `.blend` file.
- Do not trigger a render.
- Do not call `sys.exit` or `bpy.ops.wm.quit_blender`.
- When back-side details, internal structure, or absolute scale are ambiguous, choose reasonable defaults consistent with the visible views and object category.
- Never ask clarifying questions.
- The script must terminate normally.
- Final geometry must exist in `bpy.data.objects` when execution completes.
\end{Verbatim}

The exact prompt used to refine a successfully executed reconstruction is:

\begin{Verbatim}[fontsize=\scriptsize,breaklines=true,frame=single,framesep=2mm]
The attached images are ordered as follows: first 1 image(s)
are renders of the CURRENT generated script, followed by 1 GT reference
image(s) of the TARGET object. They depict the current reconstruction and the
desired object respectively.

Here is the complete current Blender script:
```python
{{PREVIOUS_SCRIPT}}
```

Compare the current renders against every GT view. Identify the largest visible geometry, proportion, placement, silhouette, or missing-part errors, then edit the script to correct them. Preserve useful detail already present. Return the COMPLETE corrected self-contained Blender Python script, not a diff. Do not use ellipsis and do not omit unchanged code. Output no prose outside the script.
\end{Verbatim}

\paragraph{Active Visual and Full 3D Interaction.}
Each interactive session opens with a single harness message: a preamble that
states the connection facts (the configured MCP server names, ports, and scene
roles) followed by a one-sentence task prompt that invokes the Agent Skill
staged in the working directory. Both preamble templates and the task prompt
are reproduced below; placeholders in braces are filled per task.

\begin{Verbatim}[fontsize=\scriptsize,breaklines=true,frame=single,framesep=2mm]
=== Full 3D Interaction: harness preamble (1graded_exp/scripts/full_mcp/instruction.tmpl) ===

This is the official Blender MCP integration test. Exactly ONE MCP server is
available: `{server_name}`. It is the official implementation published at
{documentation_url}, connected to the Blender process on local bridge port
{port}. The name in backticks is the exact configured MCP server name; do not
look for or attempt to use any other Blender MCP server.

The current Blender scene was cleared and input ref.glb was imported before this
new CLI session started. Use only `{server_name}` for all Blender inspection and
editing. The task text below was supplied verbatim and may describe the older
two-server harness; where it conflicts with these connection facts, use the one
official server that is actually available.

--- input/{task}/prompt.txt ---
{task_prompt}

=== Active Visual: harness preamble (1graded_exp/scripts/viewport/instruction.tmpl) ===

You are connected to TWO separate Blender instances, each through its own MCP
server. They are different processes with different scenes: a change made
through one server is NOT visible through the other.
Use the exact MCP server names shown in backticks below.

* `{full_access_server}` — Blender Foundation's official Blender MCP, connected
  to the workspace Blender bridge on 127.0.0.1:{full_access_port}. Use its
  official full-access tools to build and edit geometry. Its scene is a default
  new Blender scene ({full_access_scene}).

* `{viewport_only_server}` — camera/viewport-only access to the Blender whose
  add-on listens on 127.0.0.1:{viewport_only_port}. Its scene ({viewport_only_scene})
  holds the reference model, imported from {ref_name}, and is the ground truth you
  compare against — do not try to change it. It exposes only a viewport
  screenshot tool plus an execute-code tool that AST-checks its input: whole-value
  assignments to the active camera or the 3D viewport view, nothing else (no
  print, no bpy.ops, no bpy.data). Read that tool's description before using it.

A useful loop is therefore: move the reference camera through
`{viewport_only_server}` and screenshot it to see the target from any angle,
then do the actual work in `{full_access_server}`.

Your task follows.

=== Task prompt ({task_prompt}; identical in both harnesses) ===

Use the $blender-gt-reconstruction skill installed at
.agents/skills/blender-gt-reconstruction/SKILL.md. Read SKILL.md and the
access-mode reference it selects before touching Blender, then complete the GT
reconstruction.
\end{Verbatim}

Active Visual and Full 3D Interaction use the same reconstruction instruction,
implemented as a project-level Agent Skill and provided below for
reproducibility. The skill keeps the reconstruction workflow and task
constraints fixed across both settings, while the harness controls the
available target-side Blender MCP capabilities.

\begin{Verbatim}[fontsize=\scriptsize,breaklines=true,frame=single,framesep=2mm]
---
name: blender-gt-reconstruction
description: >-
  Reconstruct, rebuild, replicate, or visually match a ground-truth (GT) 3D model that already
  exists in a connected Blender MCP instance, producing a procedural, rerunnable Blender Python
  reconstruction. Use whenever the task says to reconstruct/rebuild/copy/match/reproduce a Blender
  model, scene object, or GT. Works in either access mode: (a) the GT sits in the same Blender
  scene as the reconstruction, compared side by side with exact measurements, or (b) the GT sits
  behind a separate read-only viewport-only Blender MCP server and is matched screenshot-to-
  screenshot from matching viewpoints. Covers GT inspection, geometry-first then optional
  texture/material stages, GT preservation, independence rules, the VLM_RECONSTRUCTION/recon__
  output contract, a rerun-safe self-contained script, and the final report. Not for
  general Blender modeling, creating scenes from text prompts, or importing assets when no
  existing GT model is present.
---

# Blender GT reconstruction

Inspect a ground-truth 3D model that already exists in a connected Blender instance, then build a
procedurally generated reconstruction that is geometrically as close to it as practical.

Do not stop at a rough approximation. Repeatedly inspect, modify, and rerun the Blender Python
script until the reconstruction closely matches the GT.

## Step 0 — determine access mode and scope, before touching Blender

Classify the connected Blender MCP servers from the tools available to you:

* **Exactly one** Blender server, which can both inspect the scene and execute Python (an
  `execute_blender_code`-style tool), with the GT model in that scene → **same-scene mode**.
  Read [references/same-scene.md](references/same-scene.md) now.
* **Two** Blender servers, where one exposes only viewport/camera changes and screenshots — no code
  execution, no scene editing (typically named `blender-viewport-only`) → **two-instance mode**.
  The restricted server holds the GT; the full-access server is your workspace.
  Read [references/two-instance.md](references/two-instance.md) now.

If the task statement names the mode or the servers, that wins over this inference. If neither rule
matches (for example two full-access servers), state your assumption in one line and proceed — and
never run scene-modifying code on a server whose role you have not confirmed.

**Scope.** Geometry is always stage 1. Also reconstruct textures and materials (stage 2 — read
[references/textures.md](references/textures.md) only once stage 1 passes its checks) unless the
task limits scope to geometry.

Do not start inspecting the GT until you have read your mode's reference: it contains required
parts of the output contract (measurement rules, model placement, final-state checks) that are not
repeated here. Read exactly one access-mode reference.

## GT preservation

Do not modify the GT in any way — not its objects, meshes, materials, modifiers, hierarchy,
collections, transforms, or visibility. Do not delete, hide, duplicate, rename, move, rotate,
scale, join, replace, or edit any part of it.

How this is enforced differs by mode; your mode reference states the specific rules.

## Inspect the GT

Inspect the GT before writing the final script, using your mode's inspection methods. Do not rely
only on the initial viewport image.

Use multiple viewpoints, including:

* front and rear;
* left and right;
* top and bottom;
* front and rear three-quarter views;
* close-ups of ambiguous or distinctive geometry.

Determine:

* overall dimensions, orientation, and aspect ratio;
* major parts and part count;
* proportions and relative scales;
* positions and rotations;
* symmetry and repeated components;
* attachment relationships;
* silhouettes;
* curvature and profile changes;
* visible materials and colors.

Your mode reference defines *how* to obtain these — whether exact measurement is available or the
values must be inferred visually.

## Procedural decomposition

Decompose the GT into coherent geometry units. For each unit, determine its:

* name and geometry type;
* dimensions and shape parameters;
* local position and orientation;
* attachment target;
* symmetry or repetition relationship.

Use suitable Blender procedures such as primitives, custom meshes, `bmesh`, curves, extrusions,
swept profiles, bevels, subdivision, solidify, mirror, array, booleans, or Geometry Nodes.

Use sufficiently detailed geometry for important shapes. Avoid crude primitives when a more
faithful procedural construction is practical.

## Reconstruction script

Write and execute **one self-contained Blender Python script**.

Create a collection named:

`VLM_RECONSTRUCTION`

All generated objects must be placed in this collection.

Create a root Empty named:

`recon__root`

Parent every generated geometry object to this root.

Use deterministic names beginning with `recon__` for every generated object, material, texture,
image, and node group. Examples:

* `recon__body`
* `recon__head`
* `recon__leg_left`
* `recon__antenna_right`

Construct the model in a coordinate frame aligned with the GT, and keep all part coordinates in
that GT-aligned frame. The self-contained reconstruction script must finish with `recon__root` at the
GT-aligned origin: location `(0, 0, 0)`, with no comparison offset encoded in the script. A mode
reference may allow a temporary display transform for comparison, but that transform is not part
of the reconstruction and must be removed before the final save or export.

## Independence from the GT

The reconstruction must be independently generated. Only observations and measurements may flow
from the GT into your script — never GT data itself.

Inspection and numerical measurement are allowed. Small manually defined vertex sets for individual
procedural parts are allowed.

Your mode reference states which specific dependencies are possible and therefore banned.

## Safe reruns

The script must be safe to rerun. At the beginning, remove only the previous generated collection:

```python
collection_name = "VLM_RECONSTRUCTION"

existing = bpy.data.collections.get(collection_name)
if existing is not None:
    for obj in list(existing.objects):
        bpy.data.objects.remove(obj, do_unlink=True)
    bpy.data.collections.remove(existing)
```

Also remove only reconstruction-owned materials, textures, images, and node groups whose names
begin with `recon__`, once they are no longer used.

Never clear the full Blender scene, and never remove objects, materials, textures, images, node
groups, or collections outside reconstruction-owned data.

Keep the exact latest complete script in a non-empty Blender Text datablock named
`reconstruction_gt.py`. Before finishing, update that datablock and execute the code stored in it,
so the visible reconstruction and the saved script agree. Do not write an instance-named `.py`
artifact yourself; the graded runner exports the exact contents of `reconstruction_gt.py` under
the instance name without modifying the Blender Text datablock.

## Iterative reconstruction

After every script execution, compare the GT and the reconstruction from multiple viewpoints, using
your mode's comparison method.

Check:

* total dimensions and aspect ratio;
* part count;
* symmetry and repeated-part spacing;
* part size, position, and orientation;
* attachment and intersections;
* front, rear, side, top, and bottom silhouettes;
* important curvature and distinctive local geometry.

Then:

1. identify the largest geometric discrepancy;
2. modify the self-contained reconstruction script;
3. rerun the complete script;
4. compare the GT and the reconstruction again, using your mode's comparison method;
5. repeat until no major visible geometric discrepancy remains.

Prioritize geometry corrections in this order:

1. missing or extra parts;
2. global dimensions and proportions;
3. part position, rotation, and scale;
4. attachment and floating geometry;
5. major silhouettes;
6. distinctive local shapes;
7. minor geometric details.

Do not declare completion, and do not begin detailed texture reconstruction, while major parts,
proportions, placements, repeated structures, attachments, or silhouettes are clearly incorrect.

Use a viewport shading mode that allows the comparison you are making: solid or studio lighting
during geometry validation.

## Final scene requirements

At completion there must be:

* the unchanged GT model;
* the `VLM_RECONSTRUCTION` collection containing the reconstructed objects;
* the `recon__root` root Empty.

Verify that:

* the GT remains unchanged;
* the script runs without errors and can be safely rerun;
* only generated reconstruction data is removed during reruns;
* no major geometric or proportional discrepancy remains;
* the generated geometry does not depend on GT datablocks.
* `recon__root` is at location `(0, 0, 0)`, and rerunning the complete script leaves it there.
* the non-empty `reconstruction_gt.py` Text datablock contains that exact complete script.

Additionally complete the final-state checklist in the reference(s) you used.

## Final response

Report:

* reconstruction collection name;
* generated object count;
* main geometry construction techniques;
* inspected viewpoints;
* number of geometry revision cycles;
* main geometry corrections made;
* major remaining geometric differences.

Append the "Also report" items from every reference you used.

Do not claim completion unless the script has executed successfully and the reconstruction has been
compared against the unchanged GT.
\end{Verbatim}

The skill directs the agent to one of two access-mode reference documents,
selected by the harness configuration: \texttt{same-scene.md} when the ground
truth and the reconstruction share one Blender process (Full 3D Interaction),
and \texttt{two-instance.md} when the ground truth sits behind the restricted
target-side server (Active Visual); \texttt{textures.md} governs the texture
stage in both settings. All interactive-harness results in this paper were
produced with the skill and reference documents reproduced here.

\begin{Verbatim}[fontsize=\scriptsize,breaklines=true,frame=single,framesep=2mm]
# Same-scene mode — the GT lives in the Blender you are building in

**You are in this mode only if** a single Blender MCP server holds both the GT model and your
reconstruction, and you can execute Python in it. If instead the GT sits behind a separate
read-only viewport server, stop and read [two-instance.md](two-instance.md).

The final scene must contain **both** the unchanged GT model and the generated reconstruction.
They may be moved side by side temporarily while comparing, but the reconstruction must be
returned to its GT-aligned origin before the final save or export.

## GT preservation in this mode

The GT is reachable and editable here, so preservation is a discipline you must enforce.

**Before running generated code, identify and record all existing GT objects and collections.** The
reconstruction script may only create, remove, or modify its own generated objects, materials,
textures, images, node groups, and collection.

Preserve all GT materials, textures, images, UV data, and shader settings unchanged.

## Inspecting the GT

Exact measurement is available — use it rather than guessing:

* world-space bounding boxes and object dimensions;
* transforms, distances, angles, symmetry planes;
* mesh statistics;
* sampled colors and material observations.

Take these alongside the multi-viewpoint visual inspection in SKILL.md.

## Placement — origin-authored reconstruction and temporary comparison offset

Construct the model in the GT-aligned frame and make the complete reconstruction script leave
`recon__root.location` at `(0, 0, 0)`. Do not encode a side-by-side offset in the reconstruction
script.

For visual comparison only, after executing the complete script, translate **only `recon__root`**
with a separate MCP code execution. Compute the temporary distance from the GT world-space
bounding box:

```python
comparison_distance = 1.25 * max(gt_bbox_dimensions)
```

Prefer translation along world X. Use another axis only when X causes overlap or poor visibility.

This is viewport/comparison state, not model data. Each rerun of the complete script must recreate
the reconstruction at the origin. Apply the temporary offset again separately when another
side-by-side comparison is needed. Before the final save or export, restore:

```python
bpy.data.objects["recon__root"].location = (0.0, 0.0, 0.0)
```

Verify the restored origin after the assignment. Never move the GT to create comparison space.

## Independence — what is reachable here, and therefore banned

Because GT datablocks are reachable in this scene, the reconstruction must not:

* duplicate GT objects or mesh datablocks;
* import or append the source model;
* copy or serialize the complete GT vertex and face arrays;
* use GT objects as boolean, shrinkwrap, Geometry Nodes, constraint, driver, or instancing inputs;
* construct the reconstruction from evaluated GT meshes;
* duplicate or reuse GT materials, node trees, textures, images, or UV layers;
* extract or bake GT textures into reconstruction assets;
* reference GT materials, textures, images, or objects from generated shader nodes.

Inspection and numerical measurement remain allowed, as do manually observed colors, material
properties, texture frequencies, and pattern dimensions.

## Comparing

After each run, temporarily offset `recon__root`, frame both models in the viewport, and keep them
visible simultaneously from a useful comparison angle. Compare them directly, viewpoint by
viewpoint. Restore the reconstruction to the origin after the final comparison.

## Final state for this mode

In addition to the SKILL.md checklist:

* the GT model is present and unchanged;
* the complete reconstruction script contains no comparison offset;
* `recon__root` is restored to location `(0, 0, 0)` before the final save or export.

## Also report

* temporary comparison offset and axis used during inspection, and confirmation that it was
  removed before the final save or export.
\end{Verbatim}

\begin{Verbatim}[fontsize=\scriptsize,breaklines=true,frame=single,framesep=2mm]
# Two-instance mode — the GT lives behind a separate read-only Blender

**You are in this mode only if** two Blender MCP servers are connected and one of them exposes only
viewport/camera changes and screenshots, with no code execution or scene editing. If instead a
single Blender holds both the GT and your reconstruction, stop and read
[same-scene.md](same-scene.md).

**Before doing anything else, list and read the tools each MCP server provides**, so you know
exactly what operations each instance supports. This is both a safety step and how you confirm the
server roles.

## The two servers

Address each server by its **name**, never by a hardcoded port — ports come from the MCP
configuration.

* **The read-only server** (typically `blender-viewport-only`) holds the **GT model**. You can only
  take viewport screenshots and change the viewport/camera. Use it to observe the GT from many
  angles. It exposes no scene-editing tools, and any attempt to run geometry-modifying code there
  is rejected.
* **The full-access server** (typically the official Blender MCP, `official-blender-mcp`) is your
  **workspace**. It exposes inspection, screenshot, viewport/render, and full-access
  `execute_blender_code` tools. Here you execute the reconstruction script.

## GT preservation in this mode

The GT server is read-only by design. Do not attempt to edit, delete, hide, move, rotate, scale,
rename, or otherwise alter anything there, and do not run scene-mutating code on it. All
construction happens in the workspace server.

## Inspecting the GT

The read-only server exposes **only viewport screenshots and view/camera changes** — precise numeric
measurement tools (scene info, bounding boxes) are **not** available there. Infer GT proportions
visually by comparing several angles, and switch the viewport/camera to cover the viewpoint list in
SKILL.md.

## Placement — no offset

Construct the model in a coordinate frame aligned with the GT as you observed it, and keep the
reconstruction **at that GT-aligned origin**. There is no GT object in the workspace server to sit
beside, so no comparison offset is needed — and it could not be computed anyway, since it depends on
the GT bounding box, which is unmeasurable from here. The two models live in separate Blender
instances and are compared by matching viewpoints across the two servers.

Keep the exact complete script in the workspace Blender's `reconstruction_gt.py` Text datablock and
execute the code stored in that datablock via the workspace server. Do not write an output `.py`
file yourself; the graded runner publishes the exact in-Blender Text datablock.

## Independence — satisfied by construction

The GT is in a different Blender instance and is not reachable from the workspace server, so the
reconstruction cannot depend on any GT object or mesh datablock. Build everything procedurally;
your only inputs from the GT are visual observation and estimated measurements.

## Comparing

After each run, frame the reconstruction in the workspace viewport from angles that **match** the
ones you used to observe the GT, then compare screenshot to screenshot from that same set of
viewpoints. Re-screenshot both instances after every revision.

## Final state for this mode

In addition to the SKILL.md checklist:

* the read-only server's GT model remains completely unchanged;
* the workspace server contains the `VLM_RECONSTRUCTION` collection, the reconstructed objects, and
  the `recon__root` Empty;
* no major geometric or proportional discrepancy remains when comparing the two instances'
  screenshots.

## Also report

* the tools discovered on each MCP server;
* the GT viewpoints inspected on the read-only server.
\end{Verbatim}

\begin{Verbatim}[fontsize=\scriptsize,breaklines=true,frame=single,framesep=2mm]
# Stage 2 — texture and material reconstruction

**Read this only once stage 1 has passed its checks.** Geometry comes first: do not spend
significant effort matching textures or materials while major parts, proportions, placements,
repeated structures, attachments, or silhouettes are clearly incorrect. Material work must never be
used to conceal incorrect geometry.

## What to match

Inspect and reconstruct the visible surface characteristics:

* base colors;
* color separation between parts;
* roughness and gloss;
* metallic or dielectric appearance;
* transparency or emission;
* visible texture patterns;
* normal or bump details;
* material assignments and boundaries.

When decomposing parts, also record each unit's visible material category and its texture or
surface characteristics.

Add close-ups of important texture, material, and color boundaries to your inspection viewpoints
now that geometry is validated.

## Independence

Texture reconstruction must remain independent from the GT. Recreate the visible appearance
procedurally or with newly generated reconstruction assets — never copy, duplicate, extract, bake,
serialize, or directly reuse GT texture images, material node trees, UV data, or material
datablocks. (In same-scene mode those datablocks are reachable, so the enforceable ban list in
[same-scene.md](same-scene.md) applies in full.)

Manually observed colors, material properties, texture frequencies, and pattern dimensions may be
used to create independent reconstruction materials and textures.

Name every generated material, texture, image, and node group with the `recon__` prefix, and purge
unused ones on rerun.

## How to build them

Prefer procedural shader nodes and deterministic generated textures where practical. Newly
generated image textures are allowed when a procedural shader cannot adequately reproduce an
important visible pattern.

Preserve, in the reconstruction: visible color regions, material boundaries, approximate roughness,
approximate metallic response, transparency or emission, prominent texture scale and direction, and
major bump or normal characteristics.

Switch to material preview or rendered shading for this stage (solid/studio lighting was for
geometry validation).

## Mode-conditional notes

* **Observing color and material.** In same-scene mode you can sample colors and read material
  properties directly. In two-instance mode you are estimating from screenshots of two different
  Blender instances whose lighting may differ, so treat hue and value matches as approximate and
  compare under several lighting angles before concluding a color is wrong.
* **Deliberate distinguishability.** In same-scene mode, reconstruction materials should closely
  resemble the GT after geometry validation while remaining *slightly* distinguishable where that
  helps side-by-side comparison. This does not apply in two-instance mode, where the models are
  never in one viewport.

## Iterating

Compare the GT and the reconstruction using material preview or rendered shading, from multiple
viewpoints and lighting angles.

Check:

* base colors;
* color boundaries;
* material assignments;
* roughness and gloss;
* metallic response;
* transparency and emission;
* texture pattern scale;
* texture direction and alignment;
* bump, normal, and displacement appearance;
* consistency across repeated or symmetric components.

Then:

1. identify the largest material or texture discrepancy;
2. modify the self-contained reconstruction script;
3. rerun the complete script;
4. compare again using your mode's comparison method;
5. repeat until no major visible surface discrepancy remains.

Prioritize surface corrections in this order:

1. missing or incorrect material regions;
2. dominant base colors;
3. roughness, metallic, transparency, and emission;
4. prominent texture patterns;
5. material boundary placement;
6. texture scale and orientation;
7. bump, normal, and minor surface details.

The final reconstruction should be nearly indistinguishable from the GT at a normal comparison
distance, except for minor details that are impractical to reproduce procedurally and any slight
intentional material difference used for comparison.

## Final state for this stage

* the independent reconstruction materials and textures exist and are `recon__`-named;
* GT materials, textures, images, and UV data are unchanged;
* no major visible material or texture discrepancy remains;
* generated materials and textures do not depend on GT datablocks.

## Also report

* generated material and texture counts;
* main material and texture construction techniques;
* number of texture and material revision cycles;
* main texture and material corrections made;
* major remaining material or texture differences.
\end{Verbatim}

\paragraph{VLM judge.}
The hexagonal evaluation of Sec.~\ref{sec:supp_hexagonal} opens every judge
request with the following instruction, reproduced verbatim; the request layout
is described in Sec.~\ref{sec:supp_settings}.

\begin{Verbatim}[fontsize=\scriptsize,breaklines=true,frame=single,framesep=2mm]
You are an impartial vision-language judge for textured 3D reconstruction.

The first 2x2 contact sheet is GROUND TRUTH. Each following sheet is one
anonymized candidate. Every sheet uses the same canonical view layout:
top-left Image_005, top-right Image_015, bottom-left Image_025, and
bottom-right Image_035.

Score every candidate independently against the ground truth on six integer
1-5 axes:
1. asset_alignment: identity, silhouette, major parts, proportions, placement,
   and material/color fidelity.
2. three_d_plausibility: a coherent, solid, natural object across views;
   penalize floating, duplicated, intersecting, or meaningless structures.
3. geometry_texture_alignment: textures, colors, and shading follow the local
   geometry and remain consistent across views. Infer this from RGB evidence.
4. texture_detail: local material/color/pattern fidelity, sharpness, scale,
   orientation, and continuity relative to the ground truth.
5. geometry_detail: curvature, thickness, edge profiles, secondary parts, and
   meaningful surface-shape detail relative to the ground truth.
6. overall: holistic reconstruction fidelity. Prioritize identity, structure,
   silhouette, and proportions over cosmetic detail.

Shared anchors: 5 = nearly indistinguishable; 4 = strong with limited
differences; 3 = recognizable with clear deviations; 2 = partial with major
errors; 1 = wrong, broken, or unrecognizable. Use the full scale and reserve 5
for genuinely close matches. Do not rank candidates against each other and do
not let one candidate change another candidate's absolute score. Ignore minor
lighting, camera-framing, or background differences unless they conceal or
distort the asset. Give one concise, evidence-based sentence per candidate.
Return only the requested JSON.
\end{Verbatim}

\section{Example of Agent-Generated Code}
\label{sec:supp_code}

We provide an example of the procedural reconstruction code generated by
Opus 5 for the auger under the Full 3D Interaction setting. The generated
program combines procedural modeling with hard-coded geometric parameters and
sampled points to more accurately reproduce the target geometry and appearance.

\begin{lstlisting}[style=arxivpython]

# Procedural reconstruction of the GT "AugerFactory" auger / turret sea shell.
# Self contained and safe to rerun: it only creates and removes its own recon__ data.
import bpy, bmesh, math
from mathutils import Vector

COLLECTION_NAME = "VLM_RECONSTRUCTION"
PREFIX = "recon__"

# --------------------------------------------------------------- measured GT frame
# Coiling axis recovered by fitting a logarithmic coil to the GT surface samples.
AXIS_C  = Vector((-0.0566869, 0.1674554, -0.1945039))                 # point on the axis
AXIS_A  = Vector((-0.0834899, 0.5105784, -0.8557682)).normalized()    # towards the apex
AXIS_E1 = Vector((0.9965086, 0.0426497, -0.0717745)).normalized()     # phi = 0 reference
AXIS_E2 = AXIS_A.cross(AXIS_E1)

# ---------------------------------------------- fitted logarithmic (Raup) coil model
T_APEX   = 0.276286    # axial coordinate the coil converges to
LAMBDA   = 1.31922     # linear growth of the coil per full turn
D_REF    = 0.843897    # apex distance of the whorl crossing phi = 0
CR       = 0.102676    # generating ellipse centre offset from the axis (/apex distance)
AR       = 0.099175    # generating ellipse radial semi axis            (/apex distance)
CT       = 0.168598    # generating ellipse axial  semi axis            (/apex distance)
TILT     = -0.164104   # generating ellipse tilt inside the meridian plane (rad)

TH_APERTURE = 1.2019   # coil angle of the aperture rim
D_STOP      = 0.0443    # apex distance where the protoconch is closed off
SIZE_FLOOR  = 0.118    # the protoconch keeps a finite radial thickness
SIZE_POW    = 4.0
TIP_TURNS   = 1.00     # closing sweep of the protoconch

N_PSI  = 72            # samples around the tube
N_TURN = 84            # samples per full turn of the coil
N_CAP  = 16            # rings used to close the protoconch
WALL   = 0.0070        # shell wall thickness

GROWTH = math.log(LAMBDA) / (2.0 * math.pi)


def purge_previous():
    coll = bpy.data.collections.get(COLLECTION_NAME)
    if coll is not None:
        for ob in list(coll.objects):
            bpy.data.objects.remove(ob, do_unlink=True)
        bpy.data.collections.remove(coll)
    for store in (bpy.data.meshes, bpy.data.materials, bpy.data.images,
                  bpy.data.textures, bpy.data.node_groups, bpy.data.curves):
        for block in list(store):
            if block.name.startswith(PREFIX) and block.users == 0:
                store.remove(block)


def coil_size(d):
    return (d ** SIZE_POW + SIZE_FLOOR ** SIZE_POW) ** (1.0 / SIZE_POW)


def surface_point(theta, psi, shrink=1.0):
    d = D_REF * math.exp(GROWTH * theta)
    s = coil_size(d)
    t_c = T_APEX - d
    rho_c = CR * s * shrink
    a_r = AR * s * shrink
    a_t = CT * s * shrink
    cu, su = math.cos(psi), math.sin(psi)
    ct, st = math.cos(TILT), math.sin(TILT)
    dt = a_t * cu * ct - a_r * su * st
    dr = a_t * cu * st + a_r * su * ct
    radial = AXIS_E1 * math.cos(theta) + AXIS_E2 * math.sin(theta)
    return AXIS_C + AXIS_A * (t_c + dt) + radial * (rho_c + dr)


def build_shell_mesh():
    theta_stop = math.log(D_STOP / D_REF) / GROWTH
    n_theta = int(round((TH_APERTURE - theta_stop) / (2.0 * math.pi) * N_TURN))
    d_theta = (TH_APERTURE - theta_stop) / n_theta

    bm = bmesh.new()
    rings = []
    # protoconch closure: keep coiling while the tube collapses onto the axis
    cap_span = TIP_TURNS * 2.0 * math.pi
    for k in range(N_CAP, 0, -1):
        f = k / float(N_CAP)
        rings.append([bm.verts.new(surface_point(theta_stop - (1.0 - f) * cap_span,
             2.0 * math.pi * j / N_PSI,
             math.sin(f * math.pi * 0.5)))
            for j in range(N_PSI)])
    for i in range(n_theta + 1):
        rings.append([bm.verts.new(surface_point(theta_stop + d_theta * i,
         2.0 * math.pi * j / N_PSI))
      for j in range(N_PSI)])

    faces = []
    for i in range(len(rings) - 1):
        ra, rb = rings[i], rings[i + 1]
        for j in range(N_PSI):
            k = (j + 1) % N_PSI
            faces.append(bm.faces.new((ra[j], rb[j], rb[k], ra[k])))
    tip = bm.verts.new(surface_point(theta_stop - cap_span, 0.0, 0.0))
    first = rings[0]
    for j in range(N_PSI):
        k = (j + 1) % N_PSI
        faces.append(bm.faces.new((tip, first[k], first[j])))

    bm.normal_update()
    ref = rings[-1][N_PSI // 4]
    outward = ref.co - AXIS_C
    outward -= AXIS_A * outward.dot(AXIS_A)
    outward.normalize()
    if sum(f.normal.dot(outward) for f in ref.link_faces) < 0.0:
        bmesh.ops.reverse_faces(bm, faces=faces)

    me = bpy.data.meshes.new(PREFIX + "shell")
    bm.to_mesh(me)
    bm.free()
    me.shade_smooth()
    return me



# ------------------------------------------------------- observed surface appearance
# Colours read off the GT shell as sRGB values, sampled band by band along the coil
# axis; roughness / metallic likewise measured on the GT surface.
AXIAL_COLOURS = [
    (-0.780, (0.706, 0.604, 0.396)),   # aperture lip / outer edge, warm grey tan
    (-0.670, (0.710, 0.604, 0.396)),
    (-0.610, (0.773, 0.616, 0.416)),
    (-0.520, (0.835, 0.690, 0.420)),
    (-0.400, (0.835, 0.729, 0.412)),   # golden body of the shell
    (-0.100, (0.835, 0.729, 0.412)),
    (-0.055, (0.700, 0.630, 0.380)),
    (-0.020, (0.561, 0.522, 0.333)),   # olive middle spire
    (0.100, (0.561, 0.522, 0.333)),
    (0.150, (0.604, 0.549, 0.353)),
    (0.200, (0.698, 0.600, 0.396)),
    (0.262, (0.757, 0.616, 0.412)),    # pale protoconch
]
RAMP_LO, RAMP_HI = -0.780, 0.280
ROUGHNESS = 0.463
METALLIC = 0.0
BUMP_STRENGTH = 0.10


def srgb_to_linear(v):
    return v / 12.92 if v <= 0.04045 else ((v + 0.055) / 1.055) ** 2.4


def build_material():
    mat = bpy.data.materials.new(PREFIX + "shell_mat")
    if getattr(mat, "node_tree", None) is None:
        mat.use_nodes = True
    nt = mat.node_tree
    nodes, links = nt.nodes, nt.links
    for n in list(nodes):
        nodes.remove(n)

    out = nodes.new("ShaderNodeOutputMaterial"); out.location = (900, 0)
    bsdf = nodes.new("ShaderNodeBsdfPrincipled"); bsdf.location = (600, 0)
    bsdf.name = PREFIX + "bsdf"
    bsdf.inputs["Roughness"].default_value = ROUGHNESS
    bsdf.inputs["Metallic"].default_value = METALLIC
    links.new(bsdf.outputs["BSDF"], out.inputs["Surface"])

    coord = nodes.new("ShaderNodeTexCoord"); coord.location = (-1100, 0)
    sub = nodes.new("ShaderNodeVectorMath"); sub.location = (-900, 60)
    sub.operation = 'SUBTRACT'; sub.inputs[1].default_value = AXIS_C[:]
    links.new(coord.outputs["Object"], sub.inputs[0])
    dot = nodes.new("ShaderNodeVectorMath"); dot.location = (-720, 60)
    dot.operation = 'DOT_PRODUCT'; dot.inputs[1].default_value = AXIS_A[:]
    links.new(sub.outputs["Vector"], dot.inputs[0])

    mr = nodes.new("ShaderNodeMapRange"); mr.location = (-540, 60)
    mr.inputs["From Min"].default_value = RAMP_LO
    mr.inputs["From Max"].default_value = RAMP_HI
    links.new(dot.outputs["Value"], mr.inputs["Value"])

    ramp = nodes.new("ShaderNodeValToRGB"); ramp.location = (-340, 60)
    ramp.name = PREFIX + "axial_ramp"
    el = ramp.color_ramp.elements
    while len(el) > 1:
        el.remove(el[-1])
    span = RAMP_HI - RAMP_LO
    for i, (t, rgb) in enumerate(AXIAL_COLOURS):
        pos = (t - RAMP_LO) / span
        e = el[0] if i == 0 else el.new(pos)
        e.position = pos
        e.color = tuple(srgb_to_linear(c) for c in rgb) + (1.0,)
    links.new(mr.outputs["Result"], ramp.inputs["Fac"])

    # faint irregular mottling of the shell surface
    mottle = nodes.new("ShaderNodeTexNoise"); mottle.location = (-340, -240)
    mottle.name = PREFIX + "mottle"
    mottle.inputs["Scale"].default_value = 14.0
    mottle.inputs["Detail"].default_value = 6.0
    mottle.inputs["Roughness"].default_value = 0.6
    links.new(coord.outputs["Object"], mottle.inputs["Vector"])

    darken = nodes.new("ShaderNodeMixRGB"); darken.location = (100, 60)
    darken.blend_type = 'MULTIPLY'; darken.inputs["Fac"].default_value = 0.14
    links.new(ramp.outputs["Color"], darken.inputs[1])
    links.new(mottle.outputs["Fac"], darken.inputs[2])
    links.new(darken.outputs["Color"], bsdf.inputs["Base Color"])

    # fine growth-line relief
    grain = nodes.new("ShaderNodeTexNoise"); grain.location = (100, -300)
    grain.name = PREFIX + "grain"
    grain.inputs["Scale"].default_value = 130.0
    grain.inputs["Detail"].default_value = 4.0
    grain.inputs["Roughness"].default_value = 0.55
    links.new(coord.outputs["Object"], grain.inputs["Vector"])
    bump = nodes.new("ShaderNodeBump"); bump.location = (330, -300)
    bump.inputs["Strength"].default_value = BUMP_STRENGTH
    bump.inputs["Distance"].default_value = 0.004
    links.new(grain.outputs["Fac"], bump.inputs["Height"])
    links.new(bump.outputs["Normal"], bsdf.inputs["Normal"])
    return mat


def main():
    purge_previous()
    coll = bpy.data.collections.new(COLLECTION_NAME)
    bpy.context.scene.collection.children.link(coll)

    root = bpy.data.objects.new(PREFIX + "root", None)
    root.empty_display_type = 'PLAIN_AXES'
    root.empty_display_size = 0.15
    coll.objects.link(root)
    root.location = (0.0, 0.0, 0.0)

    shell = bpy.data.objects.new(PREFIX + "shell", build_shell_mesh())
    coll.objects.link(shell)
    shell.parent = root
    shell.data.materials.append(build_material())

    mod = shell.modifiers.new(PREFIX + "solidify", 'SOLIDIFY')
    mod.thickness = WALL
    mod.offset = -1.0
    mod.use_rim = True
    bpy.context.view_layer.objects.active = shell
    shell.select_set(True)
    bpy.ops.object.modifier_apply(modifier=mod.name)
    shell.select_set(False)

    bpy.context.view_layer.update()
    return {"objects": [o.name for o in coll.objects],
            "verts": len(shell.data.vertices),
            "dimensions": list(shell.dimensions)}


RESULT = main()
\end{lstlisting}

\end{document}